\documentclass[5p,twocolumn]{elsarticle}
\usepackage{enumitem}
\usepackage{multirow}
\usepackage{subfigure}
\usepackage{algorithm}
\usepackage{algorithmic}
\usepackage{tabularx}
\usepackage{booktabs}
\usepackage{color}
\newcolumntype{C}{>{\centering\arraybackslash}X}
\newcolumntype{P}[1]{>{\centering\arraybackslash}p{#1}}
\usepackage{color}

\usepackage{amssymb}
\usepackage{amsmath}

\journal{Neurocomputing}

\begin{document}

\begin{frontmatter}



\title{Learning Fair Models without Sensitive Attributes: A Generative Approach}


\author[inst1]{Huaisheng Zhu}

\affiliation[inst1]{organization={College of Information Science and Technology},
            addressline={The Pennsylvania State University}, 
            city={University Park},
            postcode={16802}, 
            state={PA},
            country={United States}}

\author[inst1]{Enyan Dai}
\author[inst2]{Hui Liu}
\author[inst1]{Suhang Wang\corref{cor1}}
\ead{szw494@psu.edu}
\cortext[cor1]{Corresponding author. }

\affiliation[inst2]{organization={Department of Computer Science and Engineering},
            addressline={Michigan State University}, 
            city={East Lansing},
            postcode={48824}, 
            state={MI},
            country={United States}}

\begin{abstract}
Most existing fair classifiers rely on sensitive attributes to achieve fairness. However, for many scenarios, we cannot obtain sensitive attributes due to privacy and legal issues. The lack of sensitive attributes challenges many existing fair classifiers. Though we lack sensitive attributes, for many applications, there usually exists features/information of various formats that are relevant to sensitive attributes. For example, a person’s purchase history can reflect his/her race, which would help for learning fair classifiers on race. However, the work on exploring relevant features for learning fair models without sensitive attributes is rather limited. Therefore, in this paper, we study a novel problem of learning fair models without sensitive attributes by exploring relevant features. We propose a probabilistic generative framework to effectively estimate the sensitive attribute from the training data with relevant features in various formats and utilize the estimated sensitive attribute information to learn fair models. Experimental results on real-world datasets show the effectiveness of our framework in terms of both accuracy and fairness. Our source code is available at: https://github.com/huaishengzhu/FairWS.
\end{abstract}

\begin{keyword}
Fairness \sep Generative Model
\end{keyword}

\end{frontmatter}


\section{Introduction}
\label{sec:intro}

Over the past few years, machine learning models have shown great success in a wide spectrum of applications, such as credit scoring \cite{dong2021edits}, crime prediction \cite{mehrabi2021survey}, and salary prediction \cite{asuncion2007uci}. However, there is a growing concern about societal bias in training data  on demographic or sensitive attributes such as age, gender and race \cite{beutel2017data,hardt2016equality}. In particular, machine learning models trained on biased data can inherit the bias or even reinforce it. For example, a strong unfairness is found in the software COMPAS, which is used to predict the risk of a criminal to recommitting another crime~\cite{mehrabi2021survey}. It is found that COMPAS is more likely to assign a higher risk score to criminals of color even when they don't recommit another crime. Thus, bias issues in a machine learning model could cause severe fairness problems, which raises concerns about their real-world applications, especially in high-stake scenarios such as credit scoring and crime prediction.

Therefore, extensive studies have been conducted to mitigate the bias issues of machine learning models \cite{zhang2018mitigating,zafar2017fairness}, which can be generally categorized into three categories, i.e., pre-processing, in-processing, and post-processing. Pre-processing approaches process the training data to remove discrimination. For example, they can reduce the bias in the data by revising the attributes~\cite{feldman2015certifying}, generating non-discriminatory labeled data~\cite{xu2018fairgan}, and learning fair representations~\cite{locatello2019fairness}.  In-processing approaches will modify the training process of the state-of-the-art model. Typically, in-processing methods incorporate fairness constraint/regularizer into the objective function of the model, which can mitigate the bias of the models' prediction results~\cite{dwork2012fairness,zafar2017fairness}. As for post-processing algorithms,  they directly change the predictions from the trained model to meet the requirement of fairness~\cite{hardt2016equality,pleiss2017fairness}. 

Despite their effectiveness, the aforementioned approaches generally require the protected/sensitive attributes of each data sample to preprocess the data, regularize the model or post-process the predictions to achieve fairness. However, for many real-world applications, obtaining sensitive attributes is difficult due to privacy and legal issues \cite{coston2019fair,lahoti2020fairness}. For example, Consumer Financial Protection Bureau (CFBP) requires that creditors may not collect information about an applicant’s race, color, religion and other sensitive information \cite{lahoti2020fairness}.  Another scenario is the dataset collector didn't realize the potential bias issue when the dataset was built. Hence, the protected/sensitive attributes which would be useful to mitigate the bias issue of machine learning models are not collected.  The lack of sensitive attributes challenges most existing fairness-aware machine learning models as they rely on sensitive attributes to achieve fairness.
There are only very few initial works on training fair classifiers without sensitive attributes \cite{lahoti2020fairness,zhao2021you,yan2020fair}.
For example, \citeauthor{yan2020fair}~\cite{yan2020fair} introduces a clustering algorithm to obtain pseudo groups to replace the real protected groups. However, it can't guarantee that the groups they find are relevant to targeted subgroups to be protected. To resolve this problem, \citeauthor{zhao2021you}~\cite{zhao2021you}  assume that non-sensitive features that are highly correlated with sensitive attributes exist in the dataset and treat these non-sensitive features as pseudo sensitive attributes. Though effective, it is a strong assumption that these relevant features are highly correlated with sensitive attributes. 
Therefore, how to address fairness issues without knowing the sensitive attribute of each data sample is still an open problem to be addressed. 

\begin{figure}[t] 
\centering 
\includegraphics[width=0.46\textwidth]{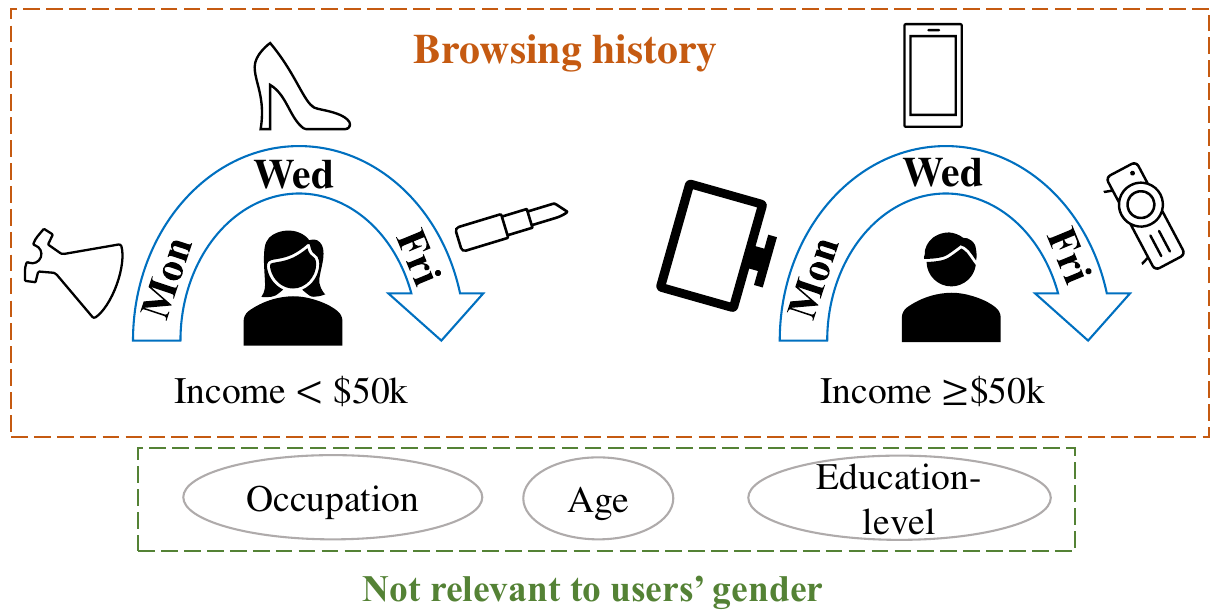} 
\vskip -1em
\caption{Illustration of irrelevant features and  relevant features (browsing history) w.r.t the sensitive attribute (gender) together with their labels (income).} 
\label{relevant}
\end{figure}

Though we lack sensitive attributes, for many real-world applications, there usually exists features/information of various formats that are not only relevant to class labels but also have a high dependency on sensitive attributes. For example,  a person’s purchase or dining history can reflect the person’s cultural background~\cite{wang2016multi}.  A user's linguistic style of reviews or browsing history can indicate the user's gender \cite{otterbacher2010inferring, phuong2014gender}. We call such  feature/information relevant to sensitive attributes as \textit{relevant features} and the remaining features as \textit{irrelevant features}. Note that those \textit{irrelevant features} might have a very weak dependency with the sensitive attributes, but would not be useful to infer sensitive attributes. Since those sensitive attributes can be in various formats, e.g., texts and graphs, they can't be directly used as pseudo-sensitive attributes. However, it is possible to estimate sensitive attributes from those relevant features, which would be useful to train a fair classifier. As the example shown in Figure~\ref{relevant}, women prefer to browse dresses, high-heeled shoes and makeup; while men tend to search electronic devices on the website. In this case, we can easily infer users' gender which is a sensitive attribute from their browsing history. In addition to browsing history, there are other attributes such as occupation, age and education-level, which are irrelevant to the user's gender but can provide useful information for the income prediction task. 

One way to alleviate bias issues is to only use those irrelevant features for prediction. However, it has several issues: (i) the labels collected might already contain bias. Such bias cannot be reduced without using sensitive attribute information to guide model learning; (ii) the relevant features are also useful for the prediction tasks. For example, items in browsing history also imply the income (the label) of the users by their price. Therefore, it's necessary to combine both browsing history and irrelevant features to train an accurate classifier. Therefore, though no sensitive attributes are provided for debiasing, it is promising to utilize relevant features to estimate sensitive attributes and adapt them to regularize the classifier for fair predictions. Meanwhile, the classifier utilizes both relevant and irrelevant features for better prediction accuracy.  However, the work on this is rather limited.

Therefore, in this paper, we study a novel problem of learning fair classifiers without sensitive attributes by estimating sensitive information from  training data. There are several challenges: (i) how can we effectively estimate sensitive information given that the sensitive attributes have a dependency on both relevant features and labels? and (ii) how to incorporate the estimated sensitive information to learn fair classifiers?
To fill this gap, we propose a novel framework \underline{Fair} Models \underline{W}ithout \underline{S}ensitive Attributes (FairWS). It adopts a probabilistic graphical model to capture the dependency between sensitive attributes, relevant features, irrelevant features and labels. Then, a Variational Autoencoder is used to model these relationships in the probabilistic graphical model, which paves a way to effectively estimate the sensitive attributes. To ensure the fairness of the given predictions, FairWS designs a fairness regularization term based on the estimated sensitive information. The main contributions are as follows: 
\begin{itemize}[leftmargin=*]
\item We study a novel problem of learning fair classifiers without sensitive attributes by estimating sensitive information from the training data; 
\item We propose a new framework FairWS, which is flexible to estimate sensitive information from  relevant features in various formats such as texts and graphs and utilize the inferred sensitive information to regularize existing classifiers to achieve fairness; 
\item We conduct experiments on real-world datasets with relevant features in various formats to show the effectiveness of FairWS for fair and high accuracy classification.
\end{itemize}

\section{Related Work}
In this section, we review related works, including fairness in machine learning and deep generative models.

\subsection{Fairness in Machine Learning}
Recent studies~\cite{feldman2015certifying,dwork2012fairness,hardt2016equality} show that machine learning models can inherit societal bias from historical data. Thus, learning fair machine learning models has attracted increasing attention and many efforts have been taken~\cite{lahoti2020fairness}, which can be generally split into three categories: (i) individual fairness~\cite{dwork2012fairness, kang2020inform, lahoti2019operationalizing,biega2018equity}, which trains the model to provide similar individuals with similar predictions; (ii) group fairness~\cite{dwork2012fairness, hardt2016equality, zhang2017achieving}, which requires the model to give equal prediction to groups with various protected sensitive attributes; (iii) Max-Min fairness~\cite{lahoti2020fairness,zhang2014fairness, hashimoto2018fairness, mohri2019agnostic}, which aims to maximize the minimum expected
utility across protected groups. We focus on group fairness in this work. 

Based on the stage of achieving fairness, existing fair machine learning methods can be split into three categories, i.e, pre-processing, in-processing, and post-processing~\cite{mehrabi2021survey}. Pre-processing methods~\cite{feldman2015certifying,xu2018fairgan,locatello2019fairness} reduce the historical discrimination in the dataset by modifying the training data. For example, \citeauthor{feldman2015certifying}~\cite{feldman2015certifying} introduce an approach to revise the attributes of training data and \citeauthor{xu2018fairgan}~\cite{xu2018fairgan} propose to generate non-discriminatory data. \citeauthor{locatello2019fairness}~\cite{locatello2019fairness} obtain fair representation for unbiased prediction.  In-processing methods~\cite{dwork2012fairness,zafar2017fairness} revise the training of fair machine learning models by designing fairness constraints or objective functions to train fair models. Post-processing methods~\cite{hardt2016equality,pleiss2017fairness} modify the prediction results from the training models to achieve fairness.

Despite their effectiveness in mitigating bias issues, the aforementioned methods require sensitive attributes of each data sample to achieve fairness, while for many scenarios, obtaining sensitive attributes is difficult due to various issues, which challenge existing fair models. Specifically, developing a fair model without sensitive attributes is crucial in real-world scenarios where obtaining sensitive attributes is important due to privacy and legal concerns \cite{coston2019fair,lahoti2020fairness}. Collecting sensitive attributes can raise ethical issues and violate privacy regulations in many domains, including health care~\cite{vokinger2021mitigating}, credit scoring~\cite{hurlin2022fairness} and criminal justice~\cite{berk2021fairness}. There is limited work on learning fair models without sensitive attributes \cite{lahoti2020fairness,zhao2021you,yan2020fair}. \citeauthor{lahoti2020fairness}~\cite{lahoti2020fairness} propose an optimization approach that leverages the notion of computationally-identifiable errors and improves the utility for worst-off protected groups. \citeauthor{yan2020fair}~\cite{yan2020fair} conducts clustering to obtain pseudo groups to substitute the real protected groups. However, the groups found by it may be irrelevant to the sensitive attributes we want the model to be fair with. For example, we might aim to make the model fair with Race but clustering gives pseudo groups for Gender. Thus, another trend of work assumes some prior knowledge about sensitive information so that their model can be fair with targeted sensitive attributes. For instance, \citeauthor{zhao2021you}~\cite{zhao2021you} assume that there are some features strongly correlated with the sensitive attributes and directly utilize these features as pseudo sensitive attributes. However, such strongly correlated features that can be treated as pseudo-sensitive attributes are not always available in real-world applications. Moreover, \citeauthor{grari2021fairness}~\cite{grari2021fairness} introduce a generative model to learn a fair model (SRCVAE) without the need for sensitive attributes, which is the most similar work to ours. The proposed approach, SRCVAE, aims to extract latent sensitive information from attributes that are influenced by sensitive attributes. However, it is important to note that attributes resulting from sensitive attributes may not exclusively contain sensitive information, as they can also incorporate non-sensitive information. Therefore, one limitation of their approach is the inability to effectively disentangle sensitive information from non-sensitive information in features that are influenced by sensitive attributes.

Our proposed FairWS is inherently different from the aforementioned approaches: (i) Instead of directly using the relevant features to obtain pseudogroups, we estimate sensitive information from relevant features to train fair and accurate classifiers. And little prior knowledge is required in FairWS to infer the sensitive information; and (ii) FairWS is flexible to learn sensitive information from relevant features of various formats, like texts and graphs. (iii) FairWS introduces a loss based on mutual information to enable the disentanglement of sensitive information from non-sensitive information in features that are influenced by sensitive attributes so that it can effectively extract relevant sensitive information from relevant features. The inferred sensitive information can be utilized to regularize existing fair models. Also, FairWS can infer sensitive information from noisy relevant features in our experiments . 

\subsection{Deep Generative Model}
Generative models aim to capture the underlying data distribution. Due to their superior performance, deep generative models like VAE~\cite{kingma2013auto} and GAN \cite{creswell2018generative} have attracted increasing attention.  \citeauthor{hu2017unifying}~\cite{hu2017unifying} provides a unified view of various deep generative models. Furthermore, there are many efforts taken to generate real data based on GANs and VAEs~\cite{hu2017toward, sohn2015learning, odena2017conditional, bowman2015generating, siarohin2018deformable}. For example, GANs are utilized to generate realistic images \cite{siarohin2018deformable}. Controlled generation of text based on VAE has been explored \cite{hu2017toward, bowman2015generating}. Recently, there are some efforts of using VAEs to resolve the fairness problem \cite{creager2019flexibly, louizos2015variational, amini2019uncovering, moyer2018invariant}. Firstly, Variational Fair Autoencoder~\cite{louizos2015variational} is proposed to  build a probabilistic  graphical model to model data with sensitive attributes. Then, a fair latent representation is obtained by removing sensitive information. \citeauthor{creager2019flexibly}~\cite{creager2019flexibly} propose to learn the latent representation of VAEs by disentangling it into two parts based on whether they are relevant to sensitive attributes. And representations irrelevant to sensitive attributes can be used as fair representation to learn fair models. \citeauthor{amini2019uncovering}~\cite{amini2019uncovering} apply fair VAEs to learn latent fair representations in facial detection systems. However, works about exploring the ability of VAEs to mitigate fairness problems without sensitive attributes are rather limited. Moreover, the proposed  FairWS is a general framework to infer sensitive information from dependency on relevant features with different formats and sensitive attributes.

\section{Problem Definition and Notations}

For many real-world applications, sensitive attributes of data samples are unavailable due to various issues such as difficulty in data collection, security or privacy issues. The lack of sensitive attributes challenges existing fairness-aware machine learning models that require sensitive attributes of data samples to achieve fairness. Though sensitive attributes for many real-world applications are unavailable, we observe that there is usually information highly relevant to sensitive attributes. For example, a person’s purchase or dining history can reflect the person's gender or cultural background, which would help learn fair classifiers on gender or cultural background. We call such information \textit{relevant features} and the remaining features that is not related to the sensitive attribute to be protected as \textit{irrelevant features}.  Note that the relevant features can be in various formats such as sequences (purchase history) and texts (reviews). Thus, the relevant features cannot be directly utilized as pseudo-sensitive attributes to regularize the model~\cite{zhao2021you}. In this paper, we aim to utilize the relevant feature to estimate sensitive attributes and train a fair and accurate classifier using both relevant and irrelevant features with estimated sensitive attributes. 

Specifically, we use $\mathcal{D}= \{\mathbf{x}_i^z, \mathbf{x}_i^r, \mathbf{y}_i\}_{i=1}^N$ to denote the training set with $N$ data samples, where $(\mathbf{x}_i^z, \mathbf{x}_i^r, \mathbf{y}_i)$ is the $i$-th sample. $\mathbf{x}_i^z$ is a feature vector that is not relevant or is very weakly related to the protected sensitive attribute $\mathbf{S}$ such as gender, $\mathbf{x}^r_i$ is relevant feature vector that are related with $\mathbf{S}$, and $\mathbf{y}_i$ is the class label. $\mathbf{x}^r_i$ can be in various formats such as purchase history and review texts. In this paper, we focus on fairness concerning a single sensitive attribute $\mathbf{S}$. We leave the extension to multi-sensitive attributes as future work. Note that we do not know the values of the sensitive attribute of each data sample. The problem is formally defined as: \\

\vspace{-0.5em}
\noindent\textit{Given the training set $\mathcal{D}_l= \{\mathbf{x}_i^z, \mathbf{x}_i^r, \mathbf{y}_i\}_{i=1}^N$ with $\mathbf{x}_i^r$ being relevant feature w.r.t protected sensitive attribute $\mathbf{S}$ and $\mathbf{x}_i^z$ being irrelevant features, we aim to learn a fair and accurate classifier $f(\mathbf{x}^z, \mathbf{x}^r) \rightarrow \hat{\mathbf{y}}$, where $f$ denotes the function to learn and $\hat{\mathbf{y}}$ represents the prediction from the classifier. And the set of predictions on test set should simultaneously maintain high accuracy and meet the fairness criteria w.r.t to the sensitive attribute $\mathbf{S}$.} 

\section{Proposed Framework}
In this section, we introduce the details of the proposed FairWS for learning fair models without sensitive attributes. Without sensitive attributes to achieve fairness, our basic idea is to estimate sensitive attributes by exploring relevant features and adopting the estimated sensitive attributes for learning fair classifiers. However, how to effectively estimate sensitive information given that the sensitive attributes have a dependency on both relevant features and labels remains a question. 
To fully capture the dependency for accurately estimating sensitive attributes, we assume that each observed data $(\mathbf{x}_i^z, \mathbf{x}_i^r, \mathbf{y}_i)$  is sampled from a probabilistic generative process which involves the latent sensitive information $\mathbf{a}_i$ and latent data representation $\mathbf{z}_i$. Thus, FairWS models the probabilistic generative process to estimate sensitive information and utilizes the estimated sensitive information to learn fair classifiers. Another important issue is about  how to incorporate the estimated sensitive information to learn fair classifiers. To resolve this problem, we introduce a regularization term to train a fair classifier with the generated latent representation of sensitive attributes. We will first introduce the probabilistic generative model for sensitive attribute estimation followed by fair classifier learning.

\subsection{Sensitive Attributes Estimation}
\label{graph}

Though the relevant features $\mathbf{x}_i^r$ are related to the sensitive attribute $\mathbf{s}_i$ of the $i$-th data sample, they cannot be simply treated as pseudo-sensitive attributes as $\mathbf{x}_i^r$ could be in various formats and might be noisy. Meanwhile, both relevant features and labels have a dependency on  sensitive attributes, which can be used to estimate the latent sensitive attributes for fair classifiers. To handle this, we use a probabilistic graphical model that models the dependency relations to obtain latent sensitive attributes. The advantages are: (i) probabilistic graphical model can capture the complex relationships among sensitive attribute $\mathbf{s}_i$, label $\mathbf{y}_i$, relevant feature $\mathbf{x}_i^r$ and irrelevant feature $\mathbf{x}_i^z$, which can help better estimate the sensitive attribute information; and (ii) estimated sensitive attribute information can be easily adopted to learn fair classifiers. 

\begin{figure}[t] 
    \centering 
    \includegraphics[width=0.2\textwidth]{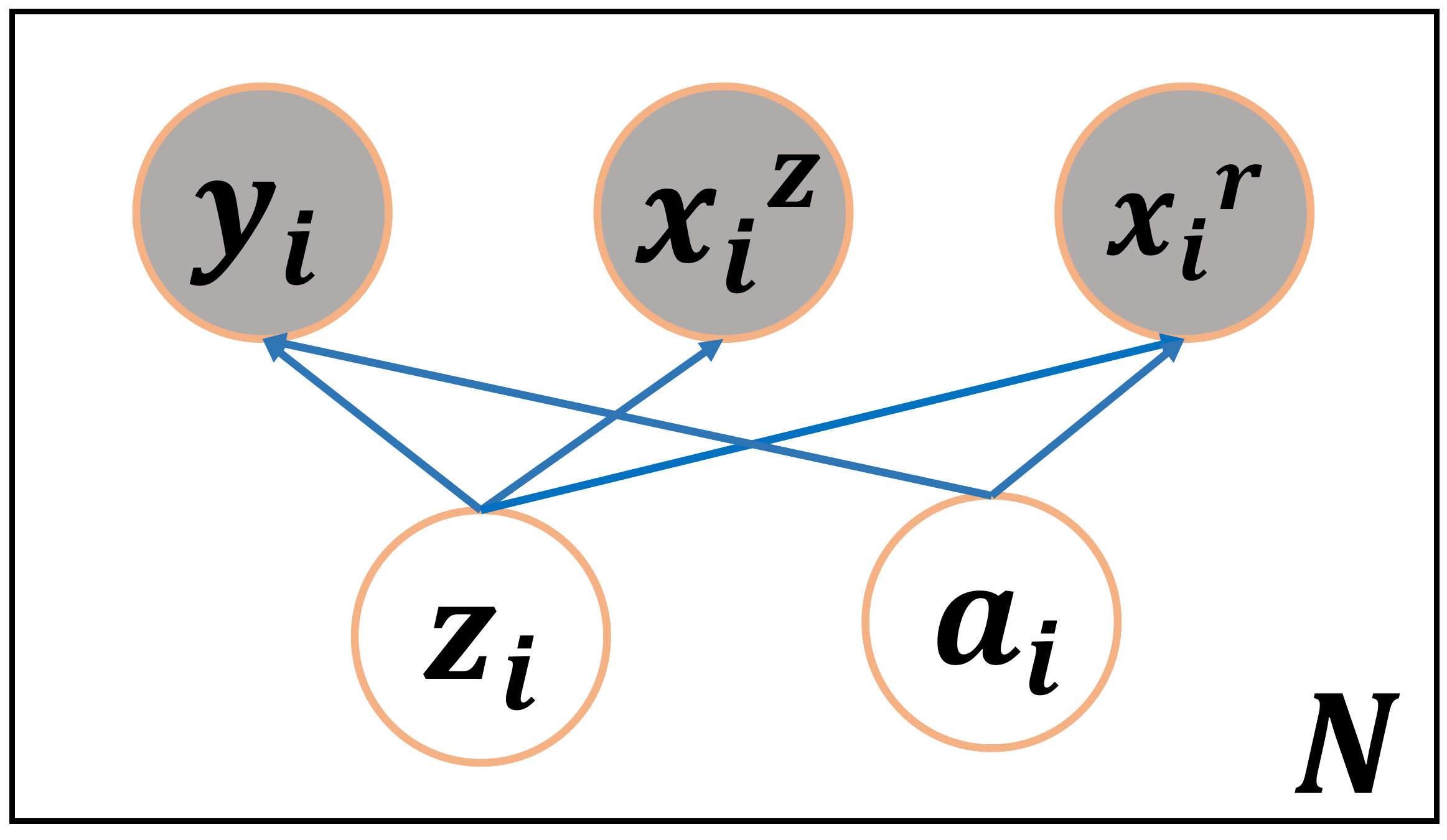} 
    \vskip -0.5em
    \caption{Probabilistic Graphical Model of FairWS. } 
    \label{Fig1}
\end{figure}

Specifically, we assume that each data sample $(\mathbf{x}_i^z, \mathbf{x}_i^r, \mathbf{y}_i)$ is sampled from a generative process as shown in Figure~\ref{Fig1}, where $\mathbf{z}_i$ is the intrinsic latent representation irrelevant to sensitive attributes $\mathbf{s}_i$, and $\mathbf{a}_i$ is the latent representation of the sensitive attribute $\mathbf{s}_i$. $\mathbf{x}_i^{r}$ is dependent on sensitive attribute's latent representation $\mathbf{a}_i$ as  $\mathbf{x}_i^{r}$ is relevant with $\mathbf{s}_i$. As the collected labels generally contain a bias towards the sensitive attribute, we assume that $\mathbf{y}_i$ is also dependent on $\mathbf{a}_i$. Since $\mathbf{z}_i$ contains the intrinsic characteristic of data sample $i$, the class label $\mathbf{y}_i$ is dependent on $\mathbf{z}_i$. $\mathbf{x}_i^r$ is also dependent on the $\mathbf{z}_i$ because $\mathbf{x}_i^r$ also contains some information that is irrelevant to sensitive attributes. It is worth noting that $\mathbf{x}_i^z$ is independent with $\mathbf{s}_i$ or has a very weak (neglectable) relationship with $\mathbf{s}_i$ because those features that are highly relevant to $\mathbf{s}_i$ are already included in $\mathbf{x}_i^r$. Hence, there is no dependency between $\mathbf{x}_i^z$ and $\mathbf{a}_i$. We disentangle $\mathbf{a}_i$ and $\mathbf{z}_i$ so that $\mathbf{a}_i$ and $\mathbf{z}_i$ can extract sensitive attribute information and non-sensitive information from $\mathbf{x}_i^r$, respectively.

\begin{figure*}[t!] 
\centering 
\includegraphics[width=0.88\textwidth]{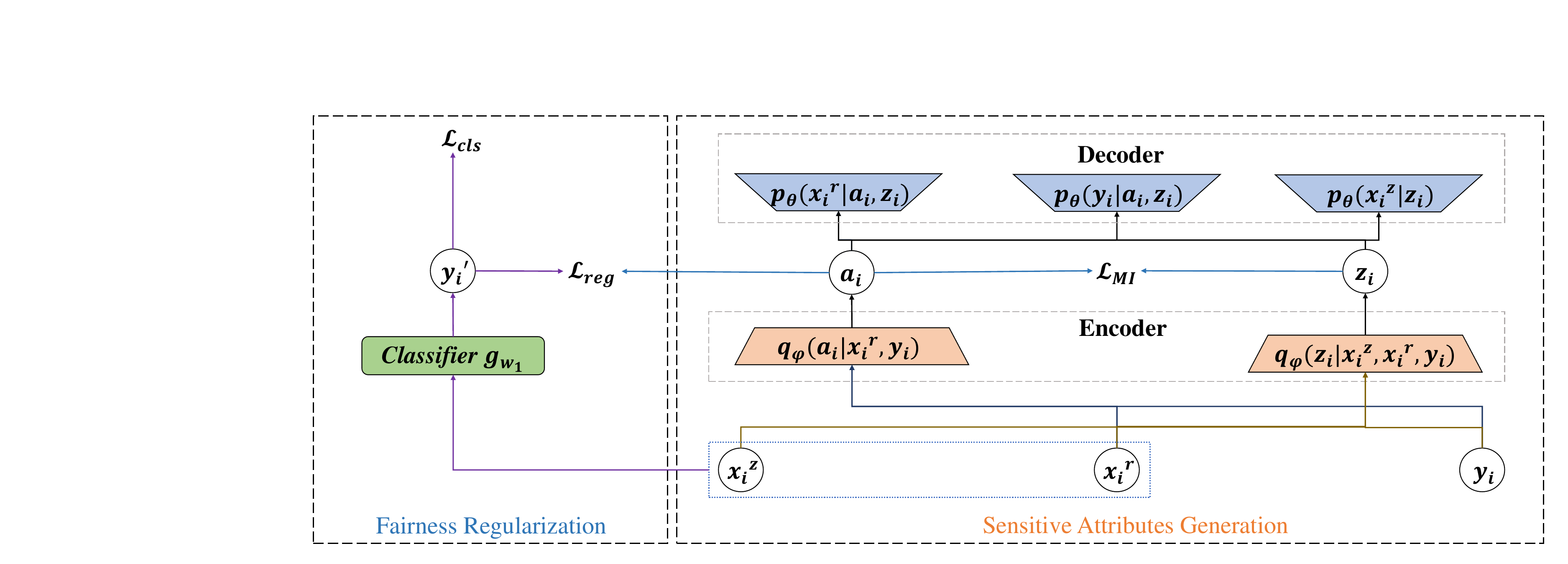} 
\flushleft 
\vskip -2em
\caption{An illustration of the proposed FairWS.} 
\label{Fig2} 
\end{figure*}

According to the probabilistic graphical model in Fig~\ref{Fig1}, $\mathbf{x}_i^z$ is independent to the latent representation of sensitive attributes $\mathbf{a}_i$, and $\mathbf{x}_i^r$ is highly correlated with $\mathbf{z}_i$. Then, the joint distribution $p(\mathbf{a}_i, \mathbf{y}_i, \mathbf{z}_i, \mathbf{x}_i^{r}, \mathbf{x}_i^{z})$ can be written as:
\begin{equation}
\begin{aligned}
    & p(\mathbf{a}_i, \mathbf{y}_i, \mathbf{z}_i, \mathbf{x}_i^{r}, \mathbf{x}_i^{z}) \\ = & 
    p(\mathbf{a}_i) p(\mathbf{z}_i) p(\mathbf{x}_i^{r} \mid \mathbf{a}_i, \mathbf{z}_i) p(\mathbf{x}_i^{z} \mid \mathbf{z}_i) p(\mathbf{y}_i \mid \mathbf{z}_i, \mathbf{a}_i),
\end{aligned}
\end{equation}
where $p(\mathbf{a}_i)$ and $p(\mathbf{z}_i)$ are the prior distributions, which are usually implemented as standard Gaussian distributions. Our goal is to maximize the likelihood of the joint distribution of observed variables, i.e., $p_{\theta}(\mathbf{x}_i^r, \mathbf{x}_i^z, \mathbf{y}_i)$. However, directly maximizing it is difficult as it contains latent variables $\mathbf{z}_i$ and $\mathbf{a}_i$. Following VAE~\cite{kingma2013auto} , we maximize the variational lower bound of this likelihood as:
\begin{equation} \label{eq:1}
\begin{aligned}
    \log p_{\theta}(\mathbf{x}_i^{r}, \mathbf{x}_i^{z}, \mathbf{y}_i) & \geq \mathbb{E}_{q_{\phi}( \mathbf{a}_i, \mathbf{z}_i \mid \mathbf{x}_i^{z}, \mathbf{x}_i^{r}, \mathbf{y}_i)}[p_{\theta}(\mathbf{x}_i^{r}, \mathbf{x}_i^{z}, \mathbf{y}_i \mid \mathbf{z}_i, \mathbf{a}_i)]\\
    &-D_{KL}(q_{\phi}(\mathbf{a}_i, \mathbf{z}_i \mid \mathbf{x}_i^{z}, \mathbf{x}_i^{r}, \mathbf{y}_i) \| p(\mathbf{z}_i, \mathbf{a}_i)),
\end{aligned}
\end{equation}
where $\mathbb{E}$ denotes the expectation,  $q_{\phi}(\mathbf{a}_i, \mathbf{z}_i \mid \mathbf{x}_i^{z}, \mathbf{x}_i^{r}, \mathbf{y}_i)$ is an auxiliary distribution to approximate $p_{\theta}(\mathbf{a}_i, \mathbf{z}_i \mid \mathbf{x}_i^{z}, \mathbf{x}_i^{r}, \mathbf{y}_i)$, and  $D_{KL}(\cdot,\cdot)$ denotes the Kullback-Leibler divergence. $\phi$ and $\theta$ are learnable parameters for neural networks of $p$ and $q$, respectively. 

As our goal is to disentangle $\mathbf{a}_i$ and $\mathbf{z}_i$, for $q_{\phi}(\mathbf{a}_i, \mathbf{z}_i \mid \mathbf{x}_i^{z}, \mathbf{x}_i^{r}, \mathbf{y}_i)$, we also assume that $\mathbf{a}_i$ and $\mathbf{z}_i$ are independent given $\mathbf{x}_i^{z}, \mathbf{x}_i^{r}, \mathbf{y}_i$, i.e.,
\begin{equation}
    q_{\phi}(\mathbf{a}_i, \mathbf{z}_i \mid \mathbf{x}_i^{z}, \mathbf{x}_i^{r}, \mathbf{y}_i) = q_{\phi}(\mathbf{a}_i \mid \mathbf{x}_i^{z}, \mathbf{y}_i)  q_{\phi}(\mathbf{z}_i \mid \mathbf{x}_i^{z}, \mathbf{x}_i^{r}, \mathbf{y}_i),
\end{equation}
where $q_{\phi}(\mathbf{a}_i \mid \mathbf{x}_i^{z}, \mathbf{y}_i)$ and $q_{\phi}(\mathbf{z}_i \mid \mathbf{x}_i^{z}, \mathbf{x}_i^{r}, \mathbf{y}_i)$ can be treated as the encoders to learn the latent representation $\mathbf{a}_i$ and $\mathbf{z}_i$ from $(\mathbf{x}_i^{z}, \mathbf{x}_i^{r}, \mathbf{y}_i)$. 

Based on Figure~\ref{Fig1}, $p_{\theta}(\mathbf{x}_i^{r}, \mathbf{x}_i^{z}, \mathbf{y}_i | \mathbf{z}_i, \mathbf{a}_i)$ can be further written as:
\begin{equation}
    p_{\theta}(\mathbf{x}_i^{r}, \mathbf{x}_i^{z}, \mathbf{y}_i | \mathbf{z}_i, \mathbf{a}_i)=p_{\theta}(\mathbf{x}_i^{r} | \mathbf{a}_i, \mathbf{z}_i) p_{\theta}(\mathbf{x}_i^{z} | \mathbf{z}_i) p_{\theta}(\mathbf{y}_i | \mathbf{z}_i, \mathbf{a}_i) \nonumber
\end{equation}
where $p_{\theta}(\mathbf{x}_{i}^{r} \mid \mathbf{a}_{i}, \mathbf{z}_{i})$, $p_{\theta}(\mathbf{y}_i \mid \mathbf{z}_i, \mathbf{a}_i)$ and
$p_{\theta}(\mathbf{x}_i^{z} \mid \mathbf{z}_i)$ are the decoders to generate $\mathbf{x}_i^r$ $\mathbf{y}_i$ and $\mathbf{x}_i^z$, respectively. The details of encoders and decoders will be discussed in section~\ref{deep}. Assume that the prior distribution $p(\mathbf{z}_i,\mathbf{a}_i)$ can be factorized as $p(\mathbf{z}_i)p(\mathbf{a}_i)$ with both $p(\mathbf{z}_i)$ and $p(\mathbf{a}_i)$ follow the normal distribution, then we can rewrite the KL divergence as:
\begin{equation}
\begin{aligned}
    &D_{KL}(q_{\phi}(\mathbf{z}_i, \mathbf{a}_i \mid \mathbf{x}_i^{z}, \mathbf{x}_i^{r}, \mathbf{y}_i) \| p(\mathbf{z}_i, \mathbf{a}_i))\\
    = & D_{KL}(q_{\phi}(\mathbf{z}_i \mid \mathbf{x}_i^{z}, \mathbf{x}_i^{r}, \mathbf{y}_i ) \| p(\mathbf{z}_i) ) +  D_{KL}(q_{\phi}(\mathbf{a}_i \mid  \mathbf{x}_i^{r}, \mathbf{y}_i ) \| p(\mathbf{a}_i) ).
\end{aligned}
\end{equation}
where $D_{KL}(q_{\phi}(\mathbf{z}_i \mid \mathbf{x}_i^{z}, \mathbf{x}_i^{r}, \mathbf{y}_i ) \| p(\mathbf{z}_i) )$ and $D_{KL}(q_{\phi}(\mathbf{a}_i \mid  \mathbf{x}_i^{r}, \mathbf{y}_i ) \| p(\mathbf{a}_i) )$ are  two KL divergence terms to regularize $q_{\phi}(\mathbf{z}_i | \mathbf{x}_i^{z}, \mathbf{x}_i^{r}, \mathbf{y}_i )$ and $q_{\phi}(\mathbf{a}_i |  \mathbf{x}_i^{r}, \mathbf{y}_i )$, respectively. 
To provide flexibility of our model, following~\cite{higgins2016beta}, we add a weight hyperparameter $\beta$ to control the influence of $D_{KL}(q_{\phi}(\mathbf{a}_i \mid  \mathbf{x}_i^{r}, \mathbf{y}_i) \| p(\mathbf{a}_i) )$. Then the variational lower bound can be written as:
\begin{equation} \label{eq:elbo_beta}
\begin{aligned}
    \mathcal{L}^i_{ELBO}&= \mathbb{E}_{q_{\phi}( \mathbf{a}_i, \mathbf{z}_i \mid \mathbf{x}_i^{z}, \mathbf{x}_i^{r}, \mathbf{y}_i)}[p_{\theta}(\mathbf{x}_i^{r}, \mathbf{x}_i^{z}, \mathbf{y}_i \mid \mathbf{z}_i, \mathbf{a}_i)] \\
    &- D_{KL}(q_{\phi}(\mathbf{z}_i \mid \mathbf{x}_i^{z}, \mathbf{x}_i^{r}, \mathbf{y}_i ) \| p(\mathbf{z}_i) ) \\
    &- \beta D_{KL}(q_{\phi}(\mathbf{a}_i \mid \mathbf{x}_i^{r}, \mathbf{y}_i ) \| p(\mathbf{a}_i) ).
\end{aligned}
\end{equation}


Since both $\mathbf{z}_i$ and $\mathbf{a}_i$ is dependent on the relevant features $\mathbf{x}^r_i$, to make sure that $\mathbf{z}_i$ and $\mathbf{a}_i$ are disentangled, i.e., $\mathbf{z}_i$ captures the class label related information from $\mathbf{x}_i^r$ while $\mathbf{z}_i$ captures sensitive attribute related information, we add a regularizer to minimize the mutual information between $\mathbf{z}_i$ and $\mathbf{a}_i$, i.e.: 
\begin{equation}
    \min_{\theta,\phi} ~ I(\mathbf{A};\mathbf{Z}) = H(\mathbf{A}) - H(\mathbf{A}\mid \mathbf{Z})
\end{equation}
where $\mathbf{A}$ is a matrix with the $i$-th row as $\mathbf{a}_i$ and $\mathbf{Z}$ is a matrix with the $i$-th row as $\mathbf{z}_i$.  $H(\cdot)$ denotes the entropy function and $I(\mathbf{A}, \mathbf{Z})$ measures dependencies between $\mathbf{A}$ and $\mathbf{Z}$. However, the mutual information $I(\mathbf{A};\mathbf{Z})$ is difficult to calculate directly. We follow \cite{belghazi2018mine} to efficiently estimate the mutual information. The basic idea is to train a neural network $Dis$ to distinguish between sample pairs from the joint distribution $p(\mathbf{a}_i,\mathbf{z}_i)$ and those from $p(\mathbf{a}_i)p(\mathbf{z}_i)$. Note that we don't need to do prior assumption on $p(\mathbf{a}_i, \mathbf{z}_i)$ and the mutual information can be approximated as:
\begin{equation} \label{eq:MI}
\begin{aligned}
    I(\mathbf{A}; \mathbf{Z}) & \approx \mathbb{E}_{p(\mathbf{a}_i, \mathbf{z}_i)}[Dis(\mathbf{a}_i, \mathbf{z}_i)]-\log \mathbb{E}_{p(\mathbf{a}_i) p(\mathbf{z}_i)}[e^{Dis(\mathbf{a}_i, \tilde{\mathbf{z}}_i)}]\\ 
    &=\mathcal{L}_{MI},
\end{aligned}
\end{equation}
where $Dis(\mathbf{a}_i,\mathbf{z}_i)$ is a binary discriminator judging if $(\mathbf{a}_i,\mathbf{z}_i)$ is from $p(\mathbf{a}_i,\mathbf{z}_i)$ or $p(\mathbf{a}_i)p(\mathbf{z}_i)$. In practice, in each batch to train $Dis$ with batch size $M$, we sample a set of $\{(\mathbf{a}_i, \mathbf{z}_i)\}_{i=1}^{M}$ from the combination of representation matrices $\mathbf{A}$ and $\mathbf{Z}$ to estimate the first term of Eq.(\ref{eq:MI}). Then, we randomly shuffle the rows of $Z$ to obtain the corrupted representation matrix $\tilde{\mathbf{Z}}$, $\{(\mathbf{a}_i, \tilde{\mathbf{z}}_i)\}_{i=1}^{M}$ is sampled from  the combination of representation matrices $\mathbf{A}$ and $\tilde{\mathbf{Z}}$ for estimating the first term of Eq.(\ref{eq:MI}).

With the ELBO in Equation~\ref{eq:elbo_beta} and the mutual information regularizer in Equation~\ref{eq:MI}, the final objective function of our sensitive attributes estimation module is:
\begin{equation}
    \min_{\theta,\phi} \frac{1}{N}\sum_{i=1}^{N} -\mathcal{L}^i_{ELBO}+ \mathcal{L}_{MI},
    \label{eq:8}
\end{equation}
Once the model is trained, we can estimate each data sample $i$'s latent representation of sensitive attributes by sampling from $q_{\phi}(\mathbf{a}_i \mid \mathbf{x}_i^{z}, \mathbf{y}_i)$. An illustration of the sensitive attribute estimation framework is shown in the right part of Fig.~\ref{Fig2}, where each term can be implemented as a neural network. To facilitate efficient large-scale training, we adopt the reparameterization trick \cite{kingma2013auto}.

\subsection{Fairness Regularization}

As shown in Figure~\ref{Fig2}, we can obtain $i$-th data sample's latent representation of sensitive attributes, $\mathbf{a}_i$, from the sensitive attributes estimation module. Then, we can use sampled $\mathbf{a}_i$ to regularize a base classifier to learn a fair classifier. One way to train a fair classifier is to only use $\mathbf{x}_i^z$. However, since both $\mathbf{x}_i^r$ and $\mathbf{x}_i^z$ contain useful information for predicting the label of the $i$-th data sample, using $\mathbf{x}_i^z$ only will lose much useful information, resulting in poor classification performance. Thus, we use both $\mathbf{x}_i^r$ and $\mathbf{x}_i^z$ to predict the label distribution of the $i$-th data sample as:
\begin{equation}
    \hat{\mathbf{y}}_i = g_{w_1}(\mathbf{x}_i^z\oplus f_{w_2}(\mathbf{x}_i^r))
    \label{eq:9}
\end{equation}
where $\oplus$ is the concatenation operator of two vectors and $g_{w_1}(\cdot)$ is multi-layered perceptrons (MLPs) to predict the label. As $\mathbf{x}_i^r$ can be in various formats such as texts and graphs, $f_{w_2}(\cdot)$ is utilized to transform $\mathbf{x}_i^r$ to a vector. For example, $f_{w_2}(\cdot)$ is the Convolutional Neural Network for text reviews and Graph Convolutional Network for graph data. $w_1$ and $w_2$ are both trainable parameters of neural networks. Note that  $f_{w_2}(\mathbf{x}_i^r)=\mathbf{x}_i^r$ if $\mathbf{x}_i^r$ is a relevant feature vector of the sample $i$. The cross-entropy loss for training the classifier $g_{w_1}$ can be written as:
\begin{equation}
    \min_{w_1, w_2} \mathcal{L}_{clf}=-\sum_{i=1}^{N} \sum_{j=1}^m \mathbf{y}_{ij} \log \hat{\mathbf{y}}_{ij},
\end{equation}
where $\mathbf{y}_{i}$ is one-hot encoding of the groundtruth label of $\mathbf{x}_i$ and $m$ is the number of class. $\hat{\mathbf{y}}_{ij}$ denotes the predicted probability of $i$-the data sampled being class $j$. 

Once well trained, the above classifier can give accurate predictions. However, $\mathbf{x}_i^r$ contains sensitive attribute information and the provided labels can also be biased, resulting in discriminatory predictions. To give fair predictions, we can utilize the estimated sensitive attributes to regularize the model. 
One way to make the prediction fair is to reduce the correlation between the prediction and the sensitive attribute $\mathbf{s}_i$. Since $\mathbf{a}_i$ is the latent representation of $\mathbf{s}_i$, the regularization form can be the correlation between predicted label vectors and latent sensitive attributes vectors as:
\begin{equation}
    \mathcal{L}_{\text {reg}}=\sum_{k=1}^{d} \sum_{j=1}^{m}|\sum_{i=1}^{N}(\hat{\mathbf{y}}_{ij}-\bar{\mathbf{y}}_{j})(\mathbf{a}_{i k}-\bar{\mathbf{a}}_{k})|,
\end{equation}
where $N$ is the number of samples and $m$ is the number of classes. $d$ is the dimension of $\mathbf{a}_i$ and $\mathbf{a}_{ik}$ is the $k$-th element of $\mathbf{a}_i$.  $\bar{\mathbf{a}}_k=\frac{1}{N}\sum_{i=1}^N \mathbf{a}_{ik}$. $\hat{\mathbf{y}}_{ij}$ denotes the predicted probability of class $j$ for sample $i$. And $\bar{\mathbf{y}}'_{j}=\frac{1}{N}\sum_{i=1}^N \mathbf{y}_{ij}$. 

The final objective function of FairWS is given as:
\begin{equation} \label{eq:11}
    \min_{w_1,w_2} \mathcal{L}_{clf} + \lambda \mathcal{L}_{reg},
\end{equation} 
where $\lambda$ is a scalar controlling the trade-off between the accuracy and fairness, and $w_1$ and $w_2$ are learnable parameters for transformation function and classifier in Eq.(\ref{eq:9}).

\subsection{Deep Learning Framework of FairWS}
\label{deep}
With the generative model for sensitive attributes given above, we will introduce the details of modeling encoders $q_{\phi}(\mathbf{z}_i | \mathbf{x}_i^{z}, \mathbf{x}_i^{r}, \mathbf{y}_i )$ and $q_{\phi}(\mathbf{a}_i |  \mathbf{x}_i^{r}, \mathbf{y}_i )$ together with decoders $p_{\theta}(\mathbf{x}_i^{r} | \mathbf{a}_i, \mathbf{z}_i)$, $p_{\theta}(\mathbf{x}_i^{z} | \mathbf{z}_i)$ and $p_{\theta}(\mathbf{y}_i | \mathbf{z}_i, \mathbf{a}_i)$ separately. In real-world applications, decoders and encoders can be very complex distributions for images and text data. In this paper, we adopt the reparameterization trick~\cite{kingma2013auto} and neural networks to model encoders and decoders, which can approximate complex distributions under mild conditions. 

Firstly, we assume the encoders $q_{\phi}(\mathbf{z}_i | \mathbf{x}_i^{z}, \mathbf{x}_i^{r}, \mathbf{y}_i )$ and $q_{\phi}(\mathbf{a}_i |  \mathbf{x}_i^{r}, \mathbf{y}_i )$ both follow the Gaussian Distribution where mean and variance are the output of the neural network. It can be defined as:
\begin{equation}
    \begin{aligned}
    q_{\phi}(\mathbf{z}_i | \mathbf{x}_i^{z}, \mathbf{x}_i^{r}, \mathbf{y}_i ) = N(\mathbf{z}_i ; \mu_{\mathbf{z}_i}, \sigma_{\mathbf{z}_i}^{2} \mathbf{I}) \quad&\mu_{\mathbf{z}_i}, \sigma_{\mathbf{z}_i}=E_z(\mathbf{x}_i^{z}, \mathbf{x}_i^{r}, \mathbf{y}_i), \\
    q_{\phi}(\mathbf{a}_i |  \mathbf{x}_i^{r}, \mathbf{y}_i ) = N(\mathbf{a}_i ; \mu_{\mathbf{a}_i}, \sigma_{\mathbf{a}_i}^{2} \mathbf{I}) \quad&\mu_{\mathbf{a}_i}, \sigma_{\mathbf{a}_i}=E_a(\mathbf{x}_i^{r}, \mathbf{y}_i),
    \end{aligned}
\end{equation}
where $\mathbf{I}$ is the identity matrix, $E_z(\cdot)$ and $E_a(\cdot)$ are the neural networks. $E_z(\cdot)$ takes $\mathbf{x}_i^{z}, \mathbf{x}_i^{r}$ and $\mathbf{y}_i$ as input and outputs the mean $\mu_{\mathbf{z}_i}$ and standard deviation $\sigma_{\mathbf{z}_i}$. Similarly, $E_a(*)$ takes  $\mathbf{x}_i^{r}$ and $\mathbf{y}_i$ as input and outputs mean $\mu_{\mathbf{a}_i}$ and variance $\sigma_{\mathbf{a}_i}$. $E_z(*)$ and $E_a(*)$ can be neural networks on the domain we are working on because $\mathbf{x}_i^{r}$ may be graph structures and text reviews. For example, for graph datasets, graph convolutional neural networks could be applied. For text datasets,  deep convolutional neural networks are good candidates. Then $\mathbf{z}_i$ and $\mathbf{a}_i$ can be sampled as $\mathbf{z}_i=\mu_{\mathbf{z}_i}+\sigma_{\mathbf{z}_i} \odot \epsilon_{z_i}$ and $\mathbf{a}_i=\mu_{\mathbf{a}_i}+\sigma_{\mathbf{a}_i} \odot \epsilon_{a_i}$, respectively, where $\epsilon_{z_i}$ and $\epsilon_{a_i}$ are random noises sampled from normal distributions.

Similarly, decoders are assumed as Guassian Distribution with mean and variance as the output of neural networks:
\begin{equation}
    \begin{aligned}
    p_{\theta}(\mathbf{x}_i^{r} | \mathbf{a}_i, \mathbf{z}_i) = N(\mathbf{x}_i^{r} ; \mu_{\mathbf{x}_i^{r}}, \sigma_{\mathbf{x}_i^{r}}^{2} \mathbf{I}) \quad&\mu_{\mathbf{x}_i^{r}} , \sigma_{\mathbf{x}_i^{r}}=D_{x^r}(\mathbf{a}_i, \mathbf{z}_i), \\
     p_{\theta}(\mathbf{x}_i^{z} | \mathbf{z}_i) = N(\mathbf{x}_i^{z} ; \mu_{\mathbf{x}_i^{z}}, \sigma_{\mathbf{x}_i^{z}}^{2} \mathbf{I}) \quad&\mu_{\mathbf{x}_i^{z}} , \sigma_{\mathbf{x}_i^{z}}=D_{x^z}(\mathbf{z}_i), \\
     p_{\theta}(\mathbf{y}_i | \mathbf{z}_i, \mathbf{a}_i) = N(\mathbf{y}_i ; \mu_{\mathbf{y}_i}, \sigma_{\mathbf{y}_i}^{2} \mathbf{I}) \quad&\mu_{\mathbf{y}_i} , \sigma_{\mathbf{y}_i}=D_{y}(\mathbf{a}_i, \mathbf{z}_i),
    \end{aligned}
\end{equation}
where $D_{x^r}(\cdot), D_{x^z}(\cdot), D_{y}(\cdot)$ are neural networks. $D_{x^r}(\cdot)$ the takes $\mathbf{a}_i$ and $\mathbf{z}_i$ as input and outputs $\mu_{\mathbf{x}_i^{r}}$ and $\sigma_{\mathbf{x}_i^{r}}^{2}$. Also, the input of of $D_{\mathbf{x}^z}(*)$ is $\mathbf{z}_i$ and output is $\mu_{\mathbf{x}_i^{z}} , \sigma_{\mathbf{x}_i^{z}}$. Then, the input of of $D_{y}(*)$ is $\mathbf{a}_i, \mathbf{z}_i$ and output is $\mu_{y_i}, \sigma_{y_i}^{2}$. And our deep learning framework is trained on Eq.(\ref{eq:8}) with mutual information loss. The overall architecture is shown in Figure~\ref{Fig2}.

\begin{algorithm}[t] 
\caption{Training Algorithm of FairWS.} 
\label{alg:Framwork} 
\begin{algorithmic}[1]
\REQUIRE
$\mathcal{D}= \{x_i^z, x_i^r, y_i\}_{i=1}^N$, $\lambda$ and $\beta$.
\ENSURE a fair classifier with $f_{w_1}$ and $g_{w_1}$

\STATE Initialize parameters of $E_z, E_a, D_{x^r}, D_{x^z}$ and $D_y$ .
\REPEAT 
\STATE 
Obtain labeled training samples $\{x_i^z, x_i^r, y_i\}_{i=1}^N$ from $\mathcal{D}$ \\
\STATE Optimized the encoder and decoder parameters $E_z$, $E_a$, $D_{x^r}$, $D_{x^z}$ by Eq.(\ref{eq:8}).
\UNTIL convergence
\STATE Infer the latent sensitive attributes $A$ = $\{a_i\}_{i=1}^N$ based on encoders and decoders
\STATE Initialize parameters of $f_{w_1}$ and $g_{w_1}$
\REPEAT 
\STATE Get all labeled samples $\{x_i^z, x_i^r, y_i\}_{i=1}^N$ from $\mathcal{D}$ and inferred senstive latent representation $A$ \\
\STATE Optimize  $f_{w_1}$ and $g_{w_1}$ by the loss from Eq.(\ref{eq:11})\\
\UNTIL convergence
\RETURN $f_{w_1}$ and $g_{w_1}$
\label{algorithm}
\end{algorithmic}
\end{algorithm}

\subsection{An Training Algorithm of FairWS}
In this subsection, we will introduce the training algorithm for FairWS. The overall process is shown in Algorithm~\ref{algorithm}. The first step of our algorithm is to generate sensitive attributes based on Graphical Probability Model in the section~\ref{graph}, which models dependency relationships between sensitive attributes, relevant features, irrelevant features and labels. Specifically, we train the encoders and decoders on labeled nodes from line 3 to line 4 on the loss Eq.(\ref{eq:8}). The implementation details are introduced in the section ~\ref{deep}. Then, the latent representation of the sensitive attributes $\mathbf{A}$ is generated via line 6. Secondly, our generate latent sensitive attributes  will be used to train fair classifiers by regularizing on $\mathbf{A}$ also with label loss as shown in Eq.(\ref{eq:11}). Finally, the output of this algorithm is a trained classifier and it will be used to predict labels on the testing set with unlabeled nodes.
\section{Experiments}

In this section, we conduct extensive experiments to evaluate the effectiveness of the proposed FairWS. Specifically, we aim to answer the following research questions:
\begin{itemize}[leftmargin=*]
    \item(\textbf{RQ1}) How does the proposed FairWS perform in terms of both classification accuracy and fairness?
    \item (\textbf{RQ2}) Can the proposed framework give accurate estimated sensitive attributes for achieving fairness?
    \item (\textbf{RQ3}) How does the quality of sensitive attributes affect the performance of the proposed FairWS?
\end{itemize}

\subsection{Datasets}
We conduct experiments on three publicly available benchmark datasets, including Adult \cite{asuncion2007uci},  Credit Defaulter \cite{dong2021edits} and Animate$\footnote{https://www.kaggle.com/marlesson/myanimelist-dataset-animes-profiles-reviews}$.

\begin{itemize}[leftmargin=*]
    \item \textbf{Adult:} This dataset contains records of personal yearly income. The task is to predict whether the yearly salary is over or under \$50,000 and the sensitive attribute is gender. It has 12 features. We use age, relation, and marital status as relevant features $\mathbf{X}^r$ and the rest as irrelevant features.
    \item \textbf{Credit Defaulter:} In this dataset, each data sample is a person which has 14 features about their personal information. In addition, two samples are connected based on the similarity of their purchase and payment records, which forms a graph.  
    The sensitive attribute of this dataset is age and the task is to classify whether a user is married. Each person's connectivity is treated as relevant features $\mathbf{X}^r$ because it is relevant to age, i.e., two persons of similar age are more likely to be connected and have similar connectivity or common friends.
    \item \textbf{Animate:} This dataset includes records of users' reviews and their profiles.  The task is to predict whether the average ranking of users' favorite movies is in the top 400 and the sensitive attribute is whether the average scores the users give to their favorite movies are above 8. Note that the ranking of movies is evaluated from the website Animate based on their popularity and rating scores. We treat the text review from users as relevant features $\mathbf{X}^r$ and their attributes as irrelevant features. Text reviews are treated as relevant features because reviews can (i) reflect people's attitudes towards the movie, i.e. whether to give this movie a higher score; and (ii) indicate their occupations, age, or other sensitive information.
\end{itemize}

The key statistics of the datasets are summarized in Table~\ref{Dataset}, which includes the number of features for each dataset, the number of class labels, the formats of their relevant features and the number of data samples. Note that  the Personal Attributes of Adult represent the feature vectors that can represent the characteristics of people. For Adult and Animate, we make the train:val:test split ratio as 5 : 2.5 : 2.5. For Credit Defaulter, following~\cite{dong2021edits}, we select 2000 nodes as the training set, 25\% for validation and 25\% for testing. Each experiment is conducted 3 times and the averaged performance will be reported.

\begin{table}[t]
    \small
    \centering
    \caption{Statistics of datasets.}
    \label{Dataset}
    \resizebox{0.95\columnwidth}{!}{
    \begin{tabular}{cccc}
        \toprule
        \textbf{Dataset} & \textbf{Adult} & \textbf{Credit Defaulter} & \textbf{Animate}\\ 
        \midrule
        Features  & 12 & 13 & 15 \\
        Class  & 2 &  2 & 2 \\
        Type of $\mathbf{X}^r$ & Personal Attributes & Graph & Text \\
        Data Size & 45,211 & 30,000 & 12,772 \\
        \bottomrule
    \end{tabular}
    }
\end{table}

\subsection{Experimental Setup}
\subsubsection{Baselines.} To evaluate the effectiveness of FairWS, we compare it with the vanilla model, sensitive-attribute-aware model and fair models without sensitive attributes.
\begin{itemize}[leftmargin=*]
    \item \textbf{Vanilla}: It utilizes the base classifier without the regularization form.  The base classifier $g_{w_1}(\cdot)$ is MLP for all datasets. The transformed model $f_{w_2}(\cdot)$ is Graph Convolutional Network (GCN) \cite{kipf2016semi} for dataset Credit Defaulter and Convolutional Neural Network (CNN) \cite{lawrence1997face} for dataset Animate.
    \item \textbf{ConstrainS}: We assume the sensitive attribute of each sample is known for this baseline. We add the correlation regularizer between sensitive attributes and the model output to the original loss of the Vanilla model. Note that ConstrainS aims to show the accuracy and fairness we can achieve, which is the upper bound for the proposed method. 
    \end{itemize}
We also include following representative models in fair learning without sensitive attributes as baselines:
\begin{itemize}[leftmargin=*]
    \item  \textbf{KSMOTE}~\cite{yan2020fair}: It conducts clustering to get pseudo groups and treats clustering groups as pseudo sensitive attributes. It then adopts fairness regularization terms with pseudo sensitive attributes to achieve fairness.
    \item  \textbf{RemoveR}~\cite{zhao2021you}: For this baseline, it removes all candidate-relevant features for fair classifiers. This baseline is utilized to validate the efficiency of regularizing classifiers with the generated latent sensitive attributes.
    \item \textbf{ConstrainR}: It trains a fair classifier by calculating the correlation regularization form on relevant features through Eq.(\ref{eq:11}). We design this baseline to show the quality of our generated latent sensitive attributes in regularizing the classifier for fair predictions.
    \item \textbf{ARL}~\cite{lahoti2020fairness}: It optimizes the model’s performance through reweighting under-represented regions detected by an adversarial model, which can alleviate bias. 
    \item  \textbf{FairRF}~\cite{zhao2021you}: It uses relevant features as pseudo sensitive attributes to regularize the model to be fair. This is the state-of-the-art method to train a fair classifier without sensitive attributes.
    \item \textbf{SRCVAE}~\cite{grari2021fairness}: It is a baseline which also focuses on training fair models without sensitive attributes. It utilizes Variational  Autoencoders to generate sensitive attributes and then uses generated sensitive attributes to learn a fair classifier. 
\end{itemize}
For baselines FairRF and ConstrainR, since the graph and text cannot be directly utilized to regularize the model, we adopt Node2vec \cite{grover2016node2vec} to obtain the node embeddings as relevant features on Credit Defaulter. Similarly,  we utilize the average of the pretrained word embeddings as relevant feature vectors on Animate for FairRF and ConstrainR.

\subsubsection{Configurations.} For ARL and KSMOTE \cite{lahoti2020fairness,yan2020fair}, we utilize the authors' source codes. For other baselines, we follow the implementation of \cite{zhao2021you}. For the decoder and encoder of our sensitive estimation module, we implement them as a multi-layer perceptron (MLP) network with two and three layers, respectively. The hidden dimension is 8 for the Adult dataset, 50 for the graph dataset, and 16 for the text dataset. For classifier $g_{w_1}(\cdot)$, we adopt three-layer MLPs for the adult dataset and one-layer for other datasets. We implement two-layer GCN and CNN for graph and text datasets separately. For fair comparison, we adopt the same backbone for all baselines. Adam optimizer is adopted to train the model, with an initial learning rate of 0.001 for all datasets.  We find the best hyperparameter $\beta$ through $\{0.001, 0.01, 0.1, 0.5, 1, 1.5\}$ and $\lambda$ through $\{0.01, 0.02, 0.03, 0.04, 0.05\}$ via grid search.

\begin{table*}
\centering
\caption{Comparison of different approaches in three datasets.}
\label{Adult_result} 
\resizebox{2.05\columnwidth}{!}{
\begin{tabular}{|c|ccc|ccc|ccc|}
\bottomrule & \multicolumn{3}{c|}{Adult} & \multicolumn{3}{c|}{Credit Defaulter} & \multicolumn{3}{c|}{ Animate} \\ 
\textbf{Methods} & \textbf{ACC} & \textbf{$\Delta_{EO}$} & \textbf{$\Delta_{DP}$} & \textbf{ACC} & \textbf{$\Delta_{EO}$} & \textbf{$\Delta_{DP}$} & \textbf{ACC} & \textbf{$\Delta_{EO}$} & \textbf{$\Delta_{DP}$} \\ \hline
Vanilla & 0.856 $\pm$ 0.001 & 0.046 $\pm$ 0.006 & 0.089 $\pm$ 0.005 & 0.731 $\pm$ 0.001 & 0.159 $\pm$ 0.001 & 0.101 $\pm$ 0.001 & 0.755 $\pm$ 0.001 & 0.330 $\pm$ 0.001 & 0.391 $\pm$ 0.001 \\
ConstrainS & 0.845  $\pm$ 0.002 & 0.040 $\pm$ 0.003 & 0.058 $\pm$ 0.004 & 0.713 $\pm$ 0.006 & 0.137 $\pm$ 0.002 & 0.087 $\pm$ 0.003 & 0.738 $\pm$ 0.008 & 0.202 $\pm$ 0.005 & 0.264 $\pm$ 0.003 \\
\hline
ARL & 0.861 $\pm$ 0.003 & 0.034 $\pm$ 0.012 & 0.141 $\pm$ 0.008 & 0.578 $\pm$ 0.001 & \textbf{0.050 $\pm$ 0.005} & \textbf{0.054 $\pm$ 0.009} & 0.688 $\pm$ 0.002 & 0.241 $\pm$ 0.003 & 0.332 $\pm$ 0.002 \\
KSMOTE & 0.560 $\pm$ 0.002 & 0.141 $\pm$ 0.031 & 0.012 $\pm$ 0.022 & 0.563 $\pm$ 0.003 & 0.203 $\pm$ 0.002 & 0.258 $\pm$ 0.001 & 0.672 $\pm$ 0.004 & 0.174 $\pm$ 0.001 & 0.320 $\pm$ 0.002 \\
RemoveR & 0.801 $\pm$ 0.010 & 0.124 $\pm$ 0.004 & 0.071 $\pm$ 0.002 & 0.674 $\pm$ 0.002 & 0.148 $\pm$ 0.003 & 0.092 $\pm$ 0.001 &  0.715 $\pm$ 0.002 & 0.193 $\pm$ 0.002 & 0.273 $\pm$ 0.002 \\
ConstrainR & 0.832 $\pm$ 0.013 & 0.061 $\pm$ 0.015 & 0.088 $\pm$ 0.019 & 0.668 $\pm$ 0.014 & 0.121 $\pm$ 0.012 & 0.089 $\pm$ 0.016 & 0.726 $\pm$ 0.017 & 0.257 $\pm$ 0.010 & 0.329 $\pm$ 0.012 \\
FairRF & 0.832 $\pm$ 0.001   & 0.025 $\pm$ 0.009 & 0.066 $\pm$ 0.004 & 0.682 $\pm$ 0.002 & 0.163 $\pm$ 0.002 & 0.106 $\pm$ 0.001 & 0.715 $\pm$ 0.002 & 0.225 $\pm$ 0.001 & 0.291 $\pm$ 0.001 \\
SRCVAE & 0.834 $\pm$ 0.002 & 0.038 $\pm$ 0.007 & 0.056 $\pm$ 0.013 & 0.712 $\pm$ 0.016 & 0.147 $\pm$ 0.009 & 0.089 $\pm$ 0.021 & 0.717 $\pm$ 0.009 & 0.188$ \pm$ 0.021 & 0.253 $\pm$ 0.017 \\
\hline
FairWS & \textbf{0.842 $\pm$ 0.004 } & 0.024 $\pm$ 0.012 & 0.054 $\pm$ 0.010 & \textbf{0.720 $\pm$ 0.012} & 0.145 $\pm$ 0.016 & 0.087 $\pm$ 0.010 & 0.726 $\pm$ 0.016 & \textbf{0.173$ \pm$ 0.014} & \textbf{0.247 $\pm$ 0.016} \\
FairWS + MI & 0.833  $\pm$ 0.006 & \textbf{0.013$ \pm$ 0.011} & \textbf{0.046 $\pm$ 0.019} & 0.719 $\pm$ 0.028 & 0.145 $\pm$ 0.019 & 0.074 $\pm$ 0.011 & \textbf{0.732 $\pm$ 0.020} & 0.178$ \pm$ 0.016 & 0.263 $\pm$ 0.012 \\
\specialrule{.08em}{0pt}{0pt}
\end{tabular}}
\end{table*}

\subsubsection{Evaluation Metrics.} For classification performance, we adopt the widely used accuracy as the evaluation metric. Following existing work on fair models~ \cite{mehrabi2021survey}, we adopt the difference in equal opportunity $\Delta_{EO}$ and demographic parity $\Delta_{DP}$ as the fairness metrics. They are defined as:
 
\textbf{Equal Opportunity} \cite{mehrabi2021survey}: Equal Opportunity requires that the model assigns the equal probability of positive instances with random protected attributes $i$, $j$ to a data point with a positive label:
\begin{equation}
    \mathbb{E}(\hat{y} \mid S=i, y=1)=\mathbb{E}(\hat{y} \mid S=j, y=1),
\end{equation}
where $\hat{y}$ is the output of the model $g_{w_2}$, $S$ represents the sensitive attribute. Note that the tasks in our experiment are  binary classification problems so $y'$ means the probability to be predicted as positive labels. In this paper, we report the difference for equal opportunity ($\Delta EO$), which is defined as:
 \begin{equation}
     \Delta_{E O}=|\mathbb{E}(y' \mid S=i, y=1)-\mathbb{E}(y' \mid S=j, y=1)|.
 \end{equation}

\textbf{Demographic Parity} \cite{mehrabi2021survey}: Demographic Parity requires that the predicted results of models are fair on different sensitive groups:
\begin{equation}
    \mathbb{E}(\hat{y} \mid S=i)=\mathbb{E}(\hat{y} \mid S=j), \forall i, j.
\end{equation}
We also report the difference in Demographic Parity:
\begin{equation}
    \Delta_{D P}=|\mathbb{E}(y' \mid S=i)-\mathbb{E}(y' \mid S=j)|
\end{equation}

Note that equal opportunity and demographic parity measure fairness from different dimensions. Equal opportunity requires similar performance across protected groups; while demographic parity focuses more on fair demographics~\cite{zhao2021you}. \textit{The smaller $\Delta_{EO}$  and  $\Delta_{DP}$ are, the more fair the model is}.

\subsection{Classification Accuracy and Fairness}
To answer \textbf{RQ1}, we conduct each experiment 3 times and report the average results along with standard deviation in terms of accuracy, $\Delta_{EO}$ and $\Delta_{DP}$ on three datasets in Table~\ref{Adult_result}. Note that FairWS + MI means that we utilize the mutual information loss for the sensitive attributes estimation module and FairWS means no mutual information loss. For all baselines, the hyperparameters are tuned via grid search on the validation dataset. From Table~\ref{Adult_result}, we make the following observations: 
\begin{itemize}[leftmargin=*]
    \item  Comparing Vanilla with ConstrainR, we can observe that 
    directly constraining relevant features $\mathbf{X}^r$ can help the model achieve fairer results on Adult, which is because $\mathbf{X}^r$ of Adult are simple features that can be easily incorporated into the covariance regularizer. However, when $\mathbf{X}^r$ is complex features such as text on Animate and graph on Credit Defaulter, ConstrainR method doesn't have too much effect, which is because the learned embedding vectors obtained from an unsupervised manner, which are treated as relevant features,  may be irrelevant to the targeted sensitive attributes. While for FairWS, it can extract the targeted sensitive information from the graph structure or text reviews flexibly via Variational Autoencoders in the sensitive attributes estimation part.
    \item Compared with baselines without the sensitive attributes, FairWS can achieve the best result in terms of the accuracy and fairness metrics on Adult datasets. Even though FairRF  can achieve fairer performance on Credit Defaulter and Animate datasets separately, it will result in a significant drop in accuracy. Also, FairRF can improve the fairness performance on the Adult dataset as shown in Table ~\ref{Adult_result}, but it can't work on the graph and text dataset. Finally, SRCVAE can achieve better performance by generating sensitive attributes compared with other baselines. However, they can't outperform our model. This is because they can't greatly eliminate the information of irrelevant features when generating sensitive attributes. Our proposed loss in Eq.(\ref{eq:MI}) can  discriminate sensitive attributes and irrelevant features so our model can have more accurate sensitive attributes. Based on our generated sensitive attributes, our model can obtain better performance. 
    \item Compared with ConstrainS which uses the grand truth of sensitive attributes, FairWS can achieve comparable performance or even better performance with a little drop in accuracy for all datasets. And our proposed mutual information loss can help the fairness regularization to get better performance on the Adult dataset. In addition, it can result in comparable results on fairness metrics with higher accuracy for the text and graph dataset. 
    \item Finally, FairWS achieve the best performance on Adult and Animate datasets in terms of accuracy and fairness metrics with different kinds of relevant features. It means that FairWS can extract sensitive information from relevant features in different formats, which proves the flexibility of our proposed model.
\end{itemize}

\subsection{Accuracy of Sensitive Attribute Estimation}
To answer \textbf{RQ2}, we conduct an experiment to show whether our generated latent representation $\mathbf{A}$ learns information about sensitive attributes. Specifically, we adopt Gaussian Mixture Model to cluster data into two clusters based on $\mathbf{A}$ for FairWS and FairWS + MI. As $\mathbf{A}$ should contain sensitive attribute information, we would expect the two clusters would correspond to two groups of different sensitive attributes. Then, we calculate the AUC score between predicted cluster and ground sensitive attributes and the results are shown in Table~\ref{main_result}, where Gaussian Mixture in the table means using Gaussian Mixture Model on raw relevant feature vectors. Note that this is an unsupervised setting and we evaluate the performance of the training set. From the table, we find that (i) we can obtain a large improvement compared with Gaussian Mixture on raw relevant features, which means that FairWS can efficiently estimate sensitive attributes via extracting sensitive information from relevant features; and (ii) our model FairWS can consistently outperform SRVAE. It verifies the effectiveness of our model to estimate sensitive attributes. Furthermore, FairWS+MI can outperform SRCVAE and FairWS on Adult and Animate datasets. It shows that mutual information loss can help $\mathbf{A}$ learn more information about sensitive attributes. It demonstrates that disentangling latent representation between $\mathbf{a}_i$ and $\mathbf{z}_i$ can help $\mathbf{a}_i$ to better learn sensitive information.

\begin{table}
\centering
\small
\caption{Comparison of different approaches for sensitive attributes estimation on three dataset.} \label{main_result}
\resizebox{1\columnwidth}{!}{
\begin{tabular}{lcccc}
               \bottomrule
               \multirow{2}{*}{Models}& \multicolumn{1}{c}{Adult}& \multicolumn{1}{c}{Credit Defaulter}& \multicolumn{1}{c}{Animate}\\
               & AUC & AUC & AUC\\
               \hline
               GM & 0.5087 $\pm$ 0.012 & 0.5238 $\pm$ 0.005 & 0.5363 $\pm$ 0.009\\
               SRCVAE &0.7021 $\pm$ 0.028 &0.6741 $\pm$ 0.019 & 0.6862 $\pm$ 0.002 \\
                FairWS &0.7481 $\pm$ 0.010 &\textbf{0.7046 $\pm$ 0.036} & 0.7804 $\pm$ 0.003 \\
                FairWS+MI &\textbf{0.7704 $\pm$ 0.046} & 0.6994 $\pm$ 0.004 & \textbf{0.7926 $\pm$ 0.001}\\

               \bottomrule
 \end{tabular}
 }
\end{table}

\subsection{Hyperparameter Sensitivity Analysis}
\begin{figure}[t]
    \centering
    \subfigure[Accuracy]{
        \begin{minipage}[t]{0.50\linewidth}
        \centering
        \vskip -2.5pt 
        \includegraphics[width=\textwidth]{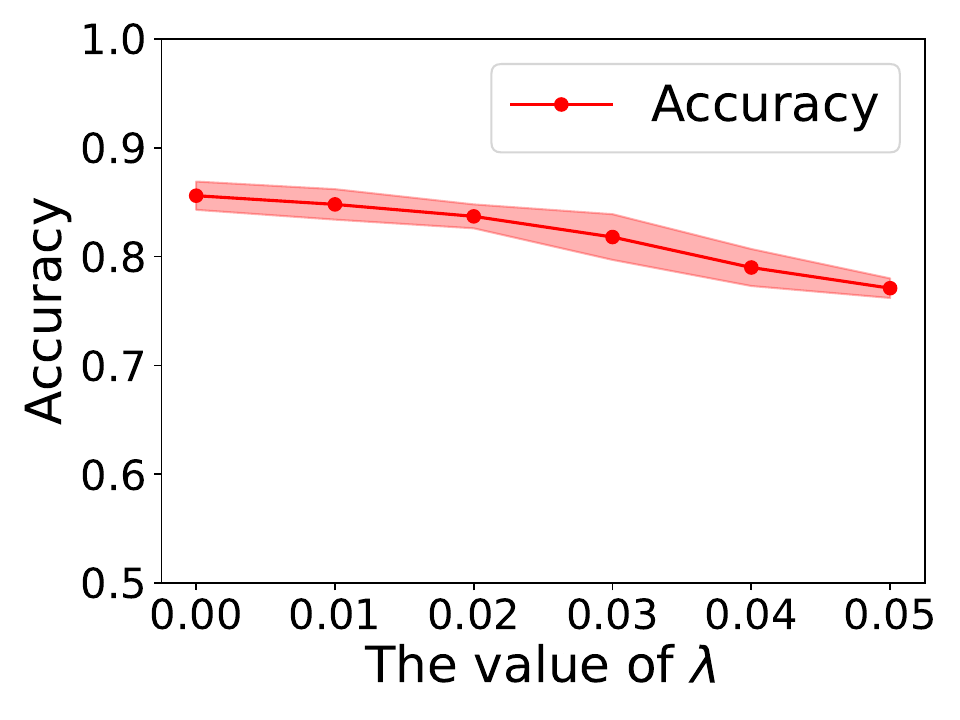}
        \end{minipage}%
    }%
    \subfigure[$\Delta_{EO}$ and $\Delta_{DP}$]{
        \begin{minipage}[t]{0.50\linewidth}
        \centering
        \vskip -2.5pt 
        \includegraphics[width=\textwidth]{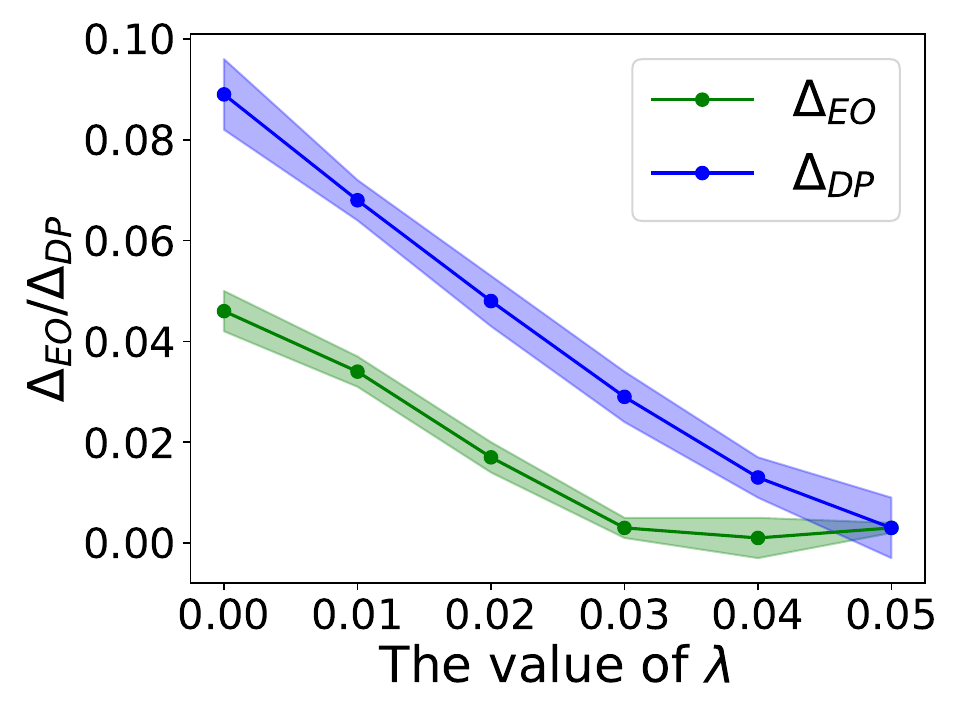}
        \end{minipage}%
    }%
    \centering
    \vspace{-4mm}
    \caption{Classification accuracy and fairness in terms of $\Delta_{EO}$ and $\Delta_{DP}$ w.r.t. the hyperparameter $\lambda$ on Adult.}
    \label{hyper_lambda_adult}
    
    \subfigure[Accuracy]{
        \begin{minipage}[t]{0.50\linewidth}
        \centering
        \vskip -2.5pt 
        \includegraphics[width=\textwidth]{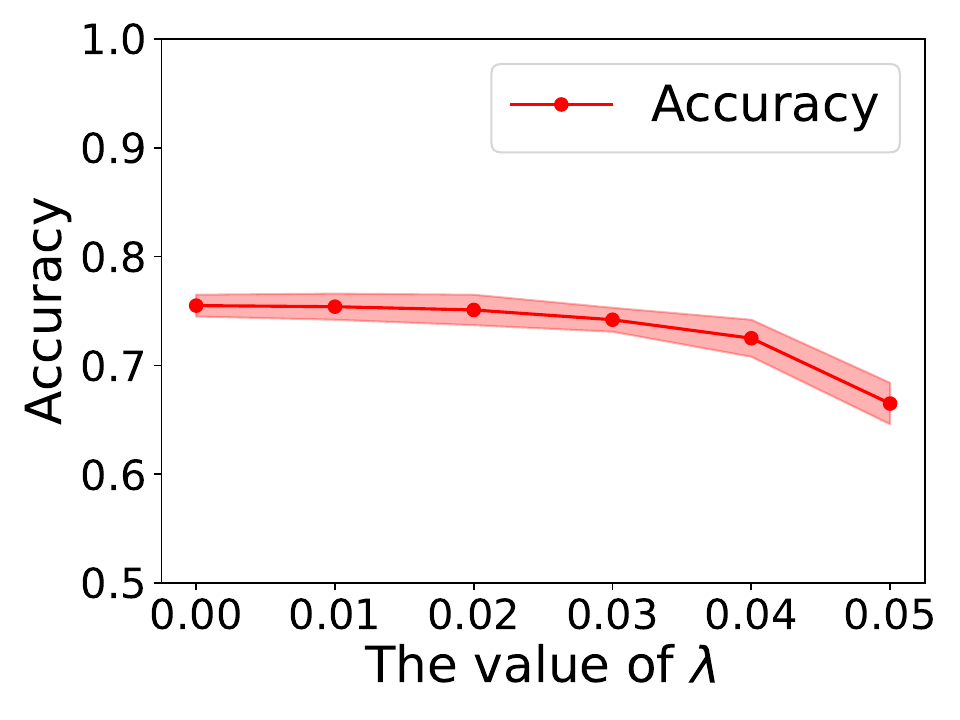}
        \end{minipage}%
    }%
    \subfigure[$\Delta_{EO}$ and $\Delta_{DP}$]{
        \begin{minipage}[t]{0.50\linewidth}
        \centering
        \vskip -2.5pt 
        \includegraphics[width=\textwidth]{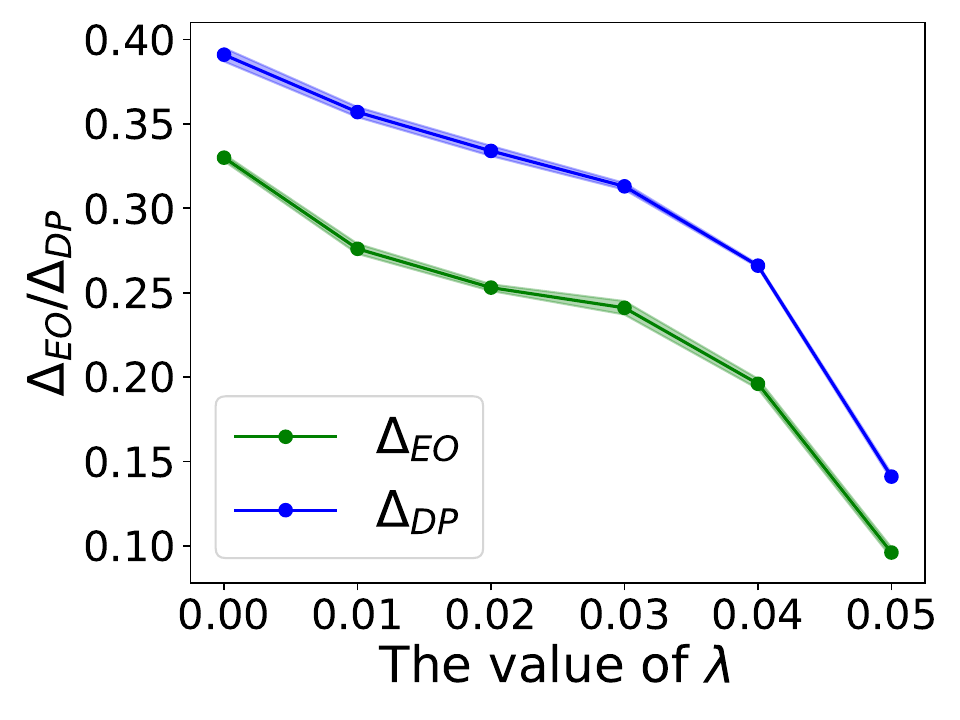}
        \end{minipage}%
    }%
    \centering
    \vspace{-4mm}
    \caption{Classification accuracy and fairness in terms of $\Delta_{EO}$ and $\Delta_{DP}$ w.r.t. the hyperparameter $\lambda$ on  Animate.}
    \vskip -1em
    \label{hyper_lambda_animate}
\end{figure}


The proposed FairWS has two important hyperparameters $\lambda$ and $\beta$. $\lambda$ controls the trade-off between fairness and accuracy when learning the fair classifier. To evaluate the parameter sensitivity on $\lambda$, we fix the sensitive attributes estimation module with $\beta=0.01$ and train a fair classifier based on the generated latent representation with different $\lambda$. We vary $\lambda$ as $\{0, 0.01, 0.02, 0.03, 0.04, 0.05\}$. 
Figure~\ref{hyper_lambda_adult} and~\ref{hyper_lambda_animate} show the results on Adult and Animate datasets separately. From Figure~\ref{hyper_lambda_adult}, we can observe that larger $\lambda$ will lead to a slight drop in terms of accuracy but significant improvement in terms of fairness $\Delta_{EO}$ and $\Delta_{DP}$, which is because the higher weight of correlation loss between predicted results and sensitive information will lead to a fairer model but with a drop of accuracy. For Figure~\ref{hyper_lambda_animate} which has text relevant features, and also shows the same pattern as Figure~\ref{hyper_lambda_adult}.   
Thus, it's important to select a $\lambda$ for various requirements, e.g., better performance on the accuracy or fairness metrics.

\begin{figure}[t]
\centering
\subfigure[Adult]{
\begin{minipage}[t]{0.50\linewidth}
\centering
\vskip -2.5pt
\includegraphics[width=\textwidth]{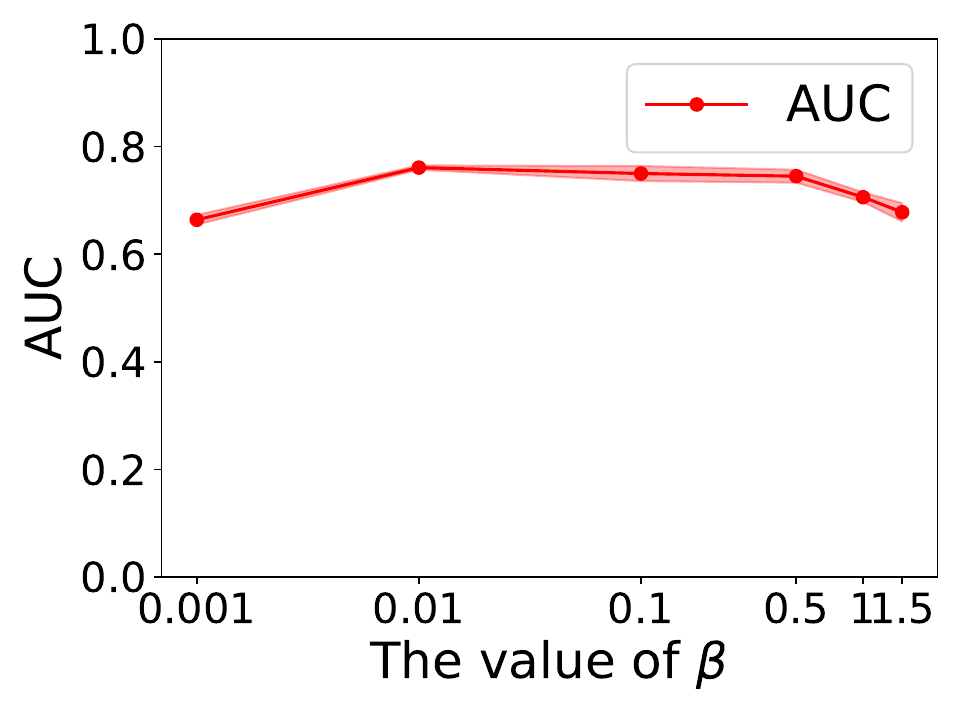}
\end{minipage}%
}%
\subfigure[Animate]{
\begin{minipage}[t]{0.50\linewidth}
\centering
\vskip -2.5pt
\includegraphics[width=\textwidth]{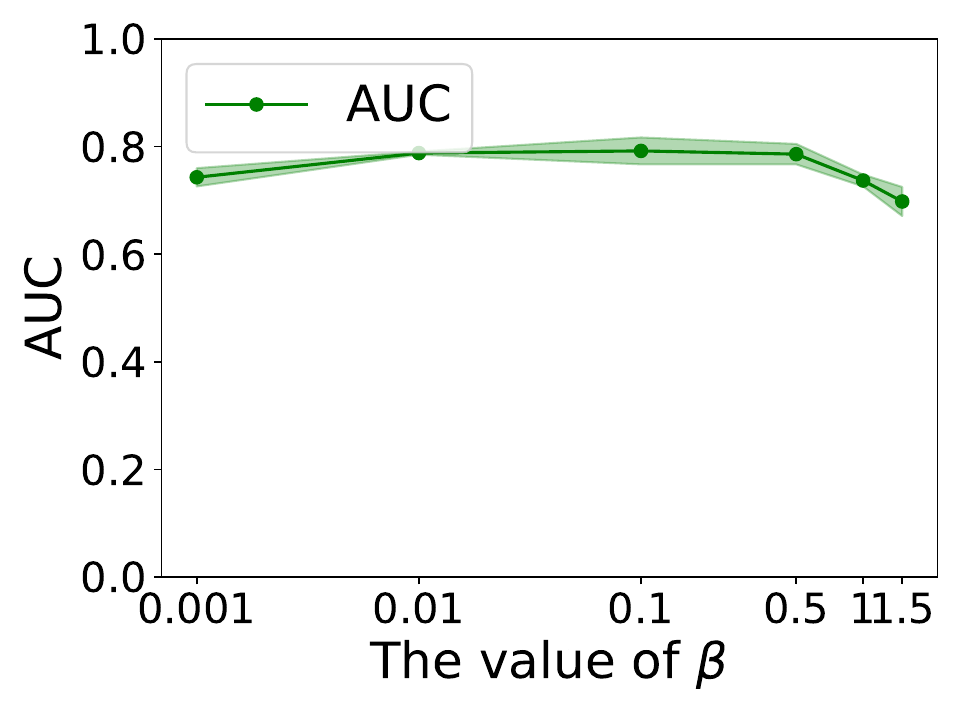}
\end{minipage}%
}%
\centering

\centering
\vspace{-4mm}
\caption{Sensitivity of $\beta$ on sensitive attribute estimation.}
\label{hyper_beta}
\vspace{-3mm}
\end{figure}

Another hyperparameter $\beta$ is to control the training process of the latent representation $\mathbf{A}$. We vary $\beta$ as $\{0.001, 0.01, 0.1, 0.5, 1, 1.5\}$. For each choice of $\beta$, we learn $\mathbf{A}$ and use Gaussian Mixture Model to predict the sensitive attributes from $\mathbf{A}$. The results of sensitive attribute estimation in terms of AUC are shown in Fig.~\ref{hyper_beta}. We observe that the performance first increases when $\beta$ increases from 0.001 to 0.01, and then results tend to be fluent on Adult and Animate. Finally, when $\beta$\ is larger than 0.5, the AUC values will have a slight drop. It is because higher $\beta$ will make $q_{\phi}(\mathbf{a}_i \mid \mathbf{x}_i^{r}, \mathbf{y}_i )$ close to standard Gaussian distributions, which will fail to extract sensitive information from $\mathbf{x}_i^{r}$.  Therefore, small and large $\beta$ will result in less sensitive information in $\mathbf{A}$ and the best value of it is between 0.01 to 0.5.

\begin{figure}[t]
\centering
\subfigure[Accuracy]{
\begin{minipage}[t]{0.50\linewidth}
\centering
\vskip -2.5pt
\includegraphics[width=\textwidth]{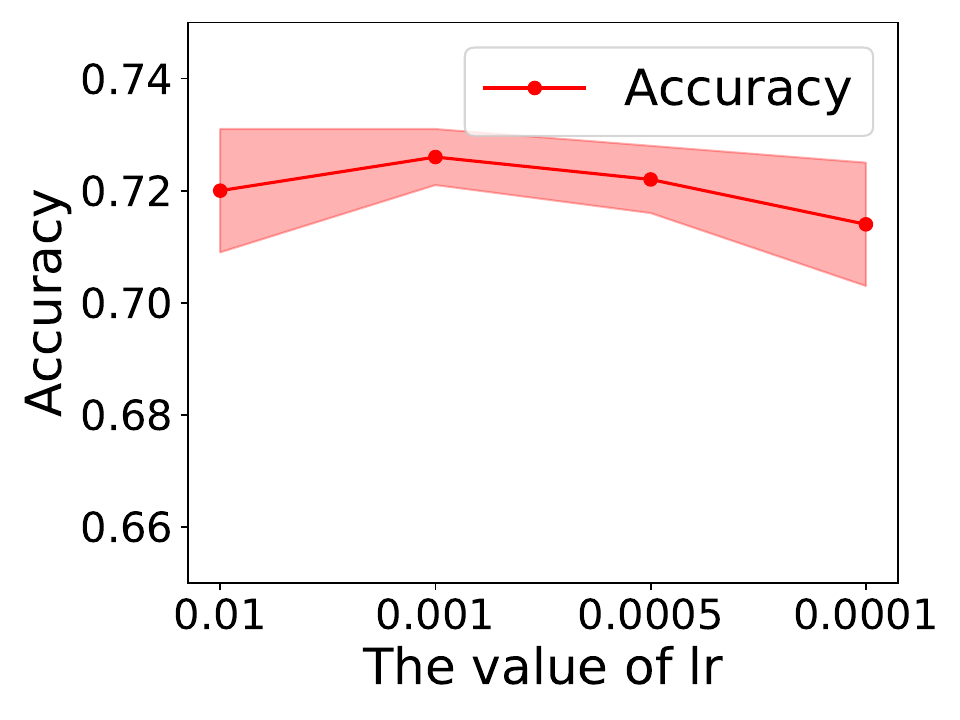}
\end{minipage}%
}%
\subfigure[$\Delta_{EO}$ and $\Delta_{DP}$]{
\begin{minipage}[t]{0.50\linewidth}
\centering
\vskip -2.5pt
\includegraphics[width=\textwidth]{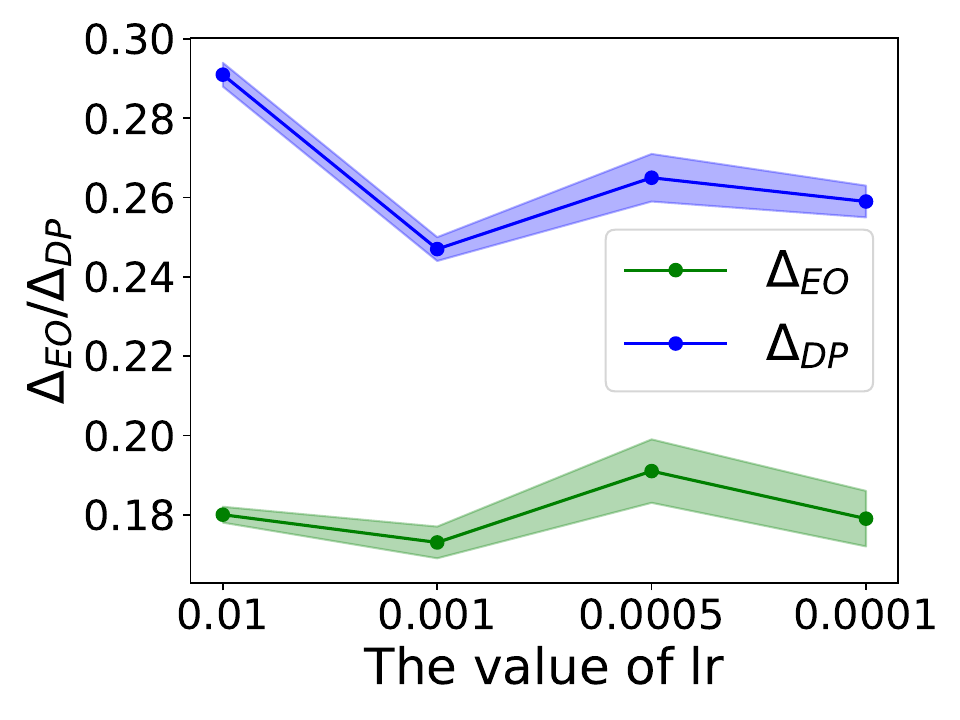}
\end{minipage}%
}%
\centering

\centering
\vspace{-4mm}
\caption{Classification accuracy and fairness in terms of $\Delta_{EO}$ and $\Delta_{DP}$ w.r.t. the hyperparameter learning rate on the Animate dataset, which is denoted as lr in the figure.}
\label{fig:lr}
\vspace{-3mm}
\end{figure}

\begin{figure}[t]
\centering
\subfigure[Accuracy]{
\begin{minipage}[t]{0.50\linewidth}
\centering
\vskip -2.5pt
\includegraphics[width=\textwidth]{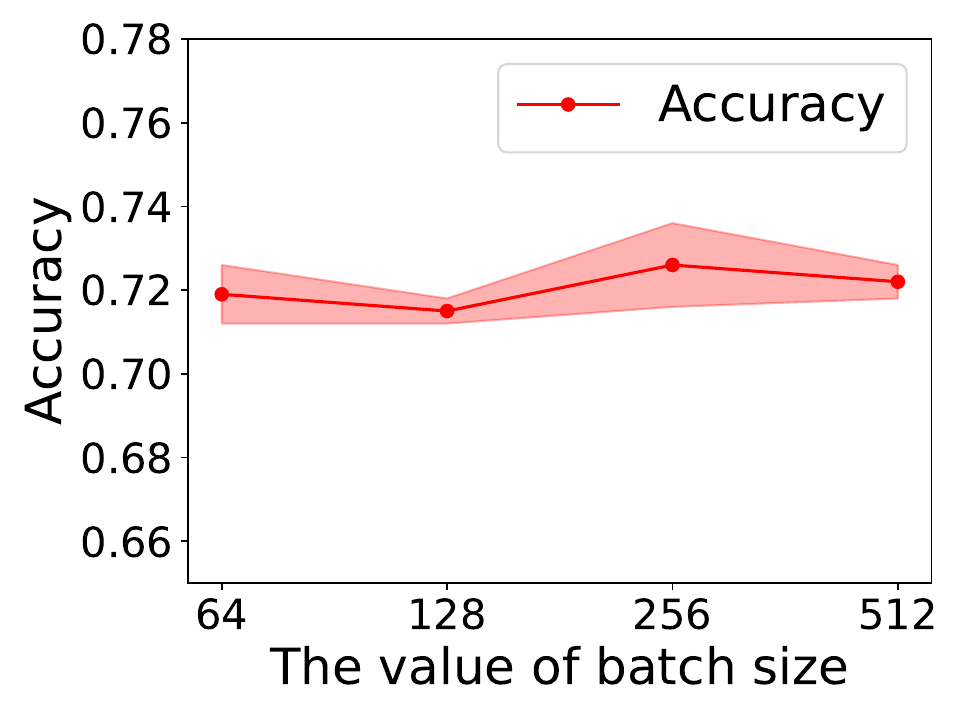}
\end{minipage}%
}%
\subfigure[$\Delta_{EO}$ and $\Delta_{DP}$]{
\begin{minipage}[t]{0.50\linewidth}
\centering
\vskip -2.5pt
\includegraphics[width=\textwidth]{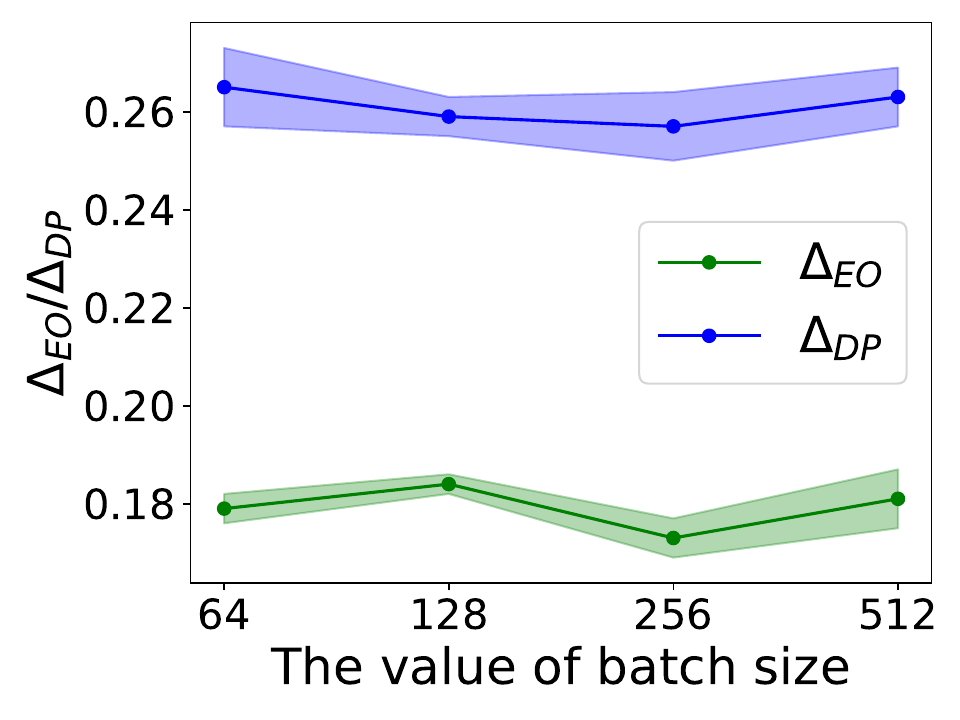}
\end{minipage}%
}%
\centering

\centering
\vspace{-4mm}
\caption{Classification accuracy and fairness in terms of $\Delta_{EO}$ and $\Delta_{DP}$ w.r.t. the hyperparameter batch size on the Animate dataset.}
\label{fig:batch}
\vspace{-3mm}
\end{figure}

We also analyze the influence of batch size and the learning rate of our model. We vary the learning rate as \{0.01, 0.001, 0.0005, 0.0001\}. Other hyperparameters are determined by cross validation with grid search. The corresponding results are shown in Figure~\ref{fig:lr}. We can observe that a higher learning rate can result in unstable training and poor performance because it causes the model to take larger steps during parameter updates. This can lead to overshooting the optimal parameter values and oscillations in the loss function. The model may fail to converge or converge to a suboptimal solution. A lower learning rate can also make the model more susceptible to getting trapped in local minima or plateaus in the loss landscape. Since the parameter updates are small, the model may struggle to escape these regions and find the global or better local optima.

We also vary the batch size as \{64, 128, 256, 512\}. Other hyperparameters are determined by cross validation with grid search. The corresponding results are shown in Figure~\ref{fig:batch}. We can find that batch size may not have too much influence on the final results because the presence of regularization techniques, such as dropout or weight decay, can further mitigate the influence of batch size on the final results by introducing robustness to the training process~\cite{srivastava2014dropout, loshchilov2017decoupled}.

\subsection{Impact of Relevant Features}

In this section, to answer \textbf{RQ3}, we explore the impact of the quality of relevant features on FiarWS. To get relevant features of different quality, we consider the following variants of FairWS:
\begin{itemize}[leftmargin=*]
    \item  \textbf{Random}: We randomly select a set of relevant features with the same number of attributes with FairWS. 
    \item \textbf{Top-1}: It includes the most-effective relevant features. We test all candidate relevant features and select the one which achieves the highest performance by regularizing a classifier based on Equation (10), and report its performance.
    \item \textbf{Noisy}: It contains features randomly selected from both highly relevant features and irrelevant features. In implementation, we replace one attribute in the highly relevant features e.g. age, relation and marital status with one irrelevant features.
\end{itemize}

\begin{table*}
\centering
\caption{The impact of the quality of relevant features on the sensitive attribute estimation  on Adult.}
\label{Impact_rel}
\resizebox{1.5\columnwidth}{!}{
    \begin{tabularx}{1\linewidth}{CCCCCC}
        \toprule
        Methods & Random & Top-1 & Noisy & GM & FairWS\\ 
        \midrule
        AUC & 0.6681$\pm$0.019 &  0.6678$\pm$0.022 & 0.7019$\pm$0.027 & 0.5078$\pm$0.018 & 0.7616$\pm$0.025 \\
        \bottomrule
    \end{tabularx}
    }
\end{table*}

We first evaluate the quality of $\mathbf{A}$ under different choices of relevant features and report the results in Table~\ref{Impact_rel}. In the table, GM means applying Gaussian Mixture on raw relevant features and we consider it as the baseline or reference result for our analysis. For a fair comparison, Random, Noisy and Top-1 are three selection methods for relevant features and we use them to select three relevant features. Then, we utilize the selected relevant features from these methods to train the FairWS model. We only conduct an experiment on Adult because Adult is the only dataset which has tabular attributes with clear semantic meaning and requires our prior knowledge to select relevant features; while the other two datasets have texts and features, which make it difficult to control the experiment. The results on Adult are shown in Table~\ref{Impact_rel}. From the table,  we can observe that FairwS can also learn information about the sensitive attributes even with noisy relevant features. Comparing FairWS with Top-1, we can find that sensitive information in one feature is limited but FairWS can utilize a set of relevant features to learn sensitive information automatically and achieve great performance.  

Furthermore, we conduct experiments to explore the impact of relevant features in terms of accuracy and fairness metrics. The results are shown in Table~\ref{rel_fair}. We make the following observation:
\begin{itemize}[leftmargin=*]
    \item  Firstly, comparing Noisy with FairWS,  with noisy relevant features where we randomly replace one highly relevant features with other features, our model can help to train a fair classifier with a little drop in accuracy. Also, in comparison with \textbf{Random}, training our models based on a random selection of features can efficiently extract sensitive information to regularize the MLP classifier. It can achieve similar results on fairness metrics with more drops in accuracy. It shows that FairRF can cope with little domain knowledge scenarios. 
    \item  In comparison with Top-1, FairWS still shows great improvement. It further proves the ability of FairWS to extract sensitive information from a set of relevant features and sensitive information in one feature is limited. 
\end{itemize}


\begin{table}
\centering
\small
\caption{Comparison of different selection approaches on relevant features.}
\label{rel_fair}

\begin{tabularx}{1\linewidth}{p{0.12\linewidth}CCC}
\bottomrule \textbf{Methods} & ACC & $\Delta_{EO}$ & $\Delta_{DP}$\\ \hline
Vanilla & 0.856 $\pm$ 0.001 & 0.046 $\pm$ 0.006 & 0.089 $\pm$ 0.005 \\
\midrule
Random & 0.826 $\pm$ 0.020 & 0.036 $\pm$ 0.015 & 0.057 $\pm$ 0.014  \\
Top-1 & 0.841 $\pm$ 0.011 & 0.041 $\pm$ 0.008 & 0.057 $\pm$ 0.010  \\
Nosiy & 0.838 $\pm$ 0.012 & 0.031 $\pm$ 0.021 & 0.059 $\pm$ 0.021  \\
\midrule
FairWS &0.842 $\pm$ 0.004  & 0.024 $\pm$ 0.012 & 0.054 $\pm$ 0.010 \\
\bottomrule
\end{tabularx}

\end{table}

\section{Conclusion}
In this paper, we study a novel problem of training fair and accurate classifiers without sensitive attributes by estimating sensitive information from features which are relevant to sensitive attributes. We propose a novel framework FairWS which learns sensitive information from relevant features and regularizes classifiers based on inferred sensitive information. FairWS can flexibly learn sensitive information from relevant features in different formats and even from noisy relevant features which may contain irrelevant features.  Through extensive experiments, we demonstrate that our method significantly outperformed the state-of-the-art methods w.r.t both accuracy and fairness metrics when sensitive attributes are unavailable. Also, we explore the impact of relevant features which proves FairWS can obtain sensitive information even with irrelevant features. Parameter sensitive analysis is also conducted to understand the sensitivity to hyperparameters. In the future, work can be done to adopt our generated sensitive information to more fair models. Furthermore, designing fair models without sensitive attributes on different kinds of data is also a promising direction, including graphs, text and images. Finally, it's also significant to explore training fair models without any prior knowledge. 

\section*{Acknowledgement}

This material is based upon work supported by, or in part by, the
National Science Foundation (NSF) under grant $\#$IIS-1909702, and
Army Research Office (ARO) under grant $\#$W911NF21-1-0198. The
findings and conclusions in this paper do not necessarily reflect the
view of the funding agency. 



\bibliographystyle{elsarticle-num-names} 
\bibliography{cas-refs}





\end{document}